%% file: main.tex
\PassOptionsToPackage{capitalize,noabbrev}{cleveref}

\documentclass[]{preprint}

\usepackage[toc,page,header]{appendix}

\usepackage[T1]{fontenc}
\usepackage{listings}
\usepackage{tcolorbox}
\tcbuselibrary{listings,breakable}
\usepackage{xcolor}
\usepackage{upquote}
\usepackage{circledsteps}

\usepackage{microtype}
\usepackage{graphicx}
\usepackage{subcaption}
\usepackage{booktabs} 

\usepackage{hyperref}

\usepackage{amsmath}
\usepackage{amssymb}
\usepackage{mathtools}
\usepackage{amsthm}

\usepackage{stmaryrd}

\theoremstyle{plain}

\theoremstyle{definition}

\theoremstyle{remark}

\usepackage{hyperref}
\usepackage{url}
\usepackage{xcolor}
\usepackage{pifont}

\usepackage[utf8]{inputenc}
\usepackage{csquotes} 
\usepackage{enumitem}
\usepackage{multirow}
\usepackage{array}
\usepackage{colortbl}

\usepackage{arydshln} 

\usepackage{wrapfig}
\usepackage{subcaption}  
\usepackage{dblfloatfix}

\usepackage[most]{tcolorbox}
\usepackage{xcolor}
\usepackage{graphicx}

\definecolor{cvprblue}{RGB}{0,102,204} 

\usepackage{amsmath,bm}

\usepackage{color}
\usepackage{tikz}
\usetikzlibrary{shapes, arrows.meta, positioning, calc, fit, backgrounds}

\usepackage{etoc}

\definecolor{sybcblue}{HTML}{004488}  
\definecolor{brickred}{HTML}{BB5566}  
\definecolor{softgray}{HTML}{666666}  

\definecolor{graybg}{gray}{0.9}

\usepackage{hyperref}
\hypersetup{
    colorlinks=true,
    linkcolor=black,
    filecolor=magenta,      
    urlcolor=seedblue, 
    pdftitle={Modality Gap},
}

\usepackage{tikz}

\usepackage{cleveref}

\definecolor{softred}{RGB}{255,230,230}
\definecolor{softblue}{RGB}{230,240,255}
\definecolor{slowcolor}{RGB}{0,100,200} 

\usepackage{amsmath}   
\usepackage{amssymb}   
\usepackage{amsthm}    

\usepackage{tcolorbox}

\usepackage{enumitem}

\usepackage{amsmath,amssymb,amsthm,mathtools}

\usepackage[utf8]{inputenc} 
\usepackage[T1]{fontenc}    
\usepackage{hyperref}       
\usepackage{url}            
\usepackage{booktabs}       
\usepackage{amsfonts}       
\usepackage{nicefrac}       
\usepackage{microtype}      
\usepackage{xcolor}         

\usepackage[utf8]{inputenc} 
\usepackage[T1]{fontenc}    
\usepackage{hyperref}       
\usepackage{url}            
\usepackage{booktabs}       
\usepackage{amsfonts}       
\usepackage{nicefrac}       
\usepackage{microtype}      
\usepackage{xcolor}         

\usepackage{tikz}
\usetikzlibrary{arrows.meta, calc, shapes.geometric, decorations.pathreplacing, positioning}

\usepackage[utf8]{inputenc} 
\usepackage[T1]{fontenc}    
\usepackage{url}            
\usepackage{booktabs}       
\usepackage{amsfonts}       
\usepackage{nicefrac}       
\usepackage{microtype}      
\usepackage{xcolor}         

\usepackage{amsmath}
\usepackage{amssymb}

\usepackage[utf8]{inputenc} 
\usepackage[T1]{fontenc}    
\usepackage{url}            
\usepackage{booktabs}       
\usepackage{amsfonts}       
\usepackage{nicefrac}       
\usepackage{microtype}      

\usepackage{wrapfig}

\usepackage{pifont}

\usepackage[table]{xcolor}
\usepackage{booktabs}
\usepackage{multirow}
\usepackage{arydshln}

\usepackage{booktabs}
\usepackage{tabularx}

\usepackage{tikz}
\usetikzlibrary{spy, arrows.meta, calc, decorations.pathreplacing, bending, shadows} 

\usepackage{amsmath,amssymb}
\usepackage{tikz}
\usetikzlibrary{
  arrows.meta,
  positioning,
  calc,
  fit,
  backgrounds,
  decorations.pathreplacing,
  shapes.geometric,
  shapes.misc,
  matrix
}

\usepackage{tikz}

\usepackage{xcolor}

\usetikzlibrary{
    arrows.meta,      
    calc,             
    shapes.geometric, 
    fadings,          
    bending           
}

\usepackage{tikz}
\usetikzlibrary{arrows.meta, calc, decorations.pathreplacing, positioning}
\usepackage{amsmath, amssymb}

\usepackage{tikz}
\usetikzlibrary{arrows.meta, calc, angles, quotes}

\usepackage{booktabs}
\usepackage{multirow}
\usepackage{graphicx}
\usepackage[table]{xcolor}
\usepackage{xspace}


\definecolor{graybg}{gray}{0.92}

\definecolor{promptblue}{HTML}{4F78B6}

\tcbset{
  promptbox/.style={
    enhanced,
    breakable,
    colframe=promptblue,
    colback=promptblue!3,
    opacityback=0.95,
    boxrule=0.7pt,
    arc=4pt,
    left=6pt,
    right=6pt,
    top=5pt,
    bottom=5pt,
    fonttitle=\bfseries,
    coltitle=white,
    colbacktitle=promptblue,
    boxed title style={
      arc=3pt,
      boxrule=0pt,
      left=5pt,
      right=5pt,
      top=2pt,
      bottom=2pt
    },
    attach boxed title to top left={
      xshift=6pt,
      yshift=-2pt
    }
  }
}

\usepackage{tikz}
\usetikzlibrary{arrows.meta, calc}
\usepackage{amsmath, amssymb}
\newcommand{\cmark}{\textcolor{green!50!black}{\ding{51}}}
\newcommand{\xmark}{\textcolor{red!75!black}{\ding{55}}}

\lstdefinestyle{prompt}{
    basicstyle=\ttfamily\scriptsize,
    breaklines=true,
    columns=fullflexible,
    frame=single,
    framerule=0.3pt,
    xleftmargin=0.6em,
    xrightmargin=0.6em,
    aboveskip=0.5em,
    belowskip=0.5em,
    showstringspaces=false
}
\usepackage{enumitem}
\usepackage{listings}
\usepackage{tcolorbox}
\tcbuselibrary{skins,breakable}
\definecolor{LightGray}{HTML}{BFBFBF}
\newtcolorbox{promptbox}[2][]{%
    enhanced,
    breakable,
    colback=gray!3,
    colframe=black!45,
    colbacktitle=gray!12,
    coltitle=black,
    fonttitle=\bfseries,
    title={#2},
    boxrule=0.5pt,
    arc=2pt,
    left=8pt,
    right=8pt,
    top=8pt,
    bottom=8pt,
    before skip=10pt,
    after skip=10pt,
    #1
}

\lstdefinestyle{prompt}{%
    basicstyle=\ttfamily\footnotesize,
    columns=fullflexible,
    keepspaces=true,
    breaklines=true,
    breakatwhitespace=false,
    showstringspaces=false,
    numbers=none,
    frame=none,
    aboveskip=6pt,
    belowskip=2pt
}

\definecolor{src}{HTML}{1B4965}
\definecolor{tgt}{HTML}{C44536}
\definecolor{gold}{HTML}{B8860B}
\definecolor{ell}{HTML}{C5D8E8}
\definecolor{gridg}{HTML}{DCDCDC}
\definecolor{note}{HTML}{6E7F8E}

\definecolor{CaseBlue}{HTML}{2471A3}

\title{Think Before You Score: Thinking Reward Model for Visual Generation}

\makeatletter

\renewcommand\author[2][]{%
  \addtolist[#1]{#2}{\authorlist}{\authorformat}{, }%
}

\makeatother

\author[1,4,*]{Xuehai Bai}
\author[2,*]{Zhenchen Tang}
\author[3,*,\spadesuit]{Yang Shi}
\author[5]{Dianyi Wang}

\makeatletter

\renewcommand\author[2][]{%
  \addtolist[#1]{#2}{\authorlist}{\authorformat}{\\[4pt]}%
}

\makeatother

\author[3]{Tengfei Liu}

\makeatletter

\renewcommand\author[2][]{%
  \addtolist[#1]{#2}{\authorlist}{\authorformat}{, }%
}

\makeatother

\author[6]{Wanshun Su}
\author[3]{Xuanyu Zhu}
\author[4]{Ruohui Wang}
\author[7]{Haiwen Diao}

\makeatletter

\renewcommand\author[2][]{%
  \addtolist[#1]{#2}{\authorlist}{\authorformat}{\\[4pt]}%
}

\makeatother

\author[8,\dagger]{Haotian Wang}

\makeatletter

\renewcommand\author[2][]{%
  \addtolist[#1]{#2}{\authorlist}{\authorformat}{, }%
}

\makeatother

\author[1,\dagger]{Xiaoling Gu}
\author[3]{Yuanxing Zhang}

\affiliation[1]{HDU}
\affiliation[2]{CASIA}
\affiliation[3]{PKU}
\affiliation[4]{SenseTime}
\affiliation[5]{FDU}
\affiliation[6]{NWPU}
\affiliation[7]{NTU}
\affiliation[8]{THU}

\newcommand{\emailaddr}[1]{%
  \href{mailto:#1}{\nolinkurl{#1}}%
}

\newcommand{\PDGRPO}{PD-GRPO\xspace}
\newcommand{\name}{TRM\xspace}

\providecommand{\firstpagefootnotes}{}

\renewcommand{\firstpagefootnotes}{%
  \parbox{0.96\textwidth}{%
    {\color{seedblue}\hrule height 0.2pt\relax}
    \vspace{6pt}
    \raggedright\normalsize
    $^{*}$ Equal Contribution. \\
    $^{\spadesuit}$ Project Lead. \\
    $^{\dagger}$ Corresponding Author. \\
  }%
}

\abstract{
Visual reward models are essential for evaluating and improving visual generation models, yet existing approaches typically map task conditions and candidate outputs directly to scalar rewards, leaving implicit what should be evaluated for each individual case.
We introduce \textbf{Think Before You Score}, a paradigm that explicitly determines \emph{what matters} for each case before judging \emph{how well} the candidate performs.
Following this principle, we propose the \textbf{Thinking Reward Model (\name)}, which formulates case-adaptive rubrics, performs rubric-guided assessment, and produces fine-grained pointwise rewards.
We further observe that conventional pairwise preference optimization can induce score polarization, and introduce \textbf{Pairwise Dual-Group Relative Policy Optimization (\PDGRPO)}, which leverages pairwise supervision to improve reward discrimination while preserving fine-grained pointwise scoring.
Extensive experiments on image generation and editing reward-modeling benchmarks demonstrate that \name{} achieves state-of-the-art performance among open-source reward models while remaining highly competitive with proprietary alternatives.
Moreover, using \name{} as a reward for reinforcement learning consistently improves diverse visual generation models, demonstrating that its fine-grained, case-adaptive rewards translate into effective optimization signals for visual generation.
}

\date{\today}

\checkdata[Project Page]{\textcolor{magenta}{\url{https://bxhsort.github.io/Thinking-Reward-Model/}}}
\checkdata[Huggingface]{\textcolor{magenta}{\url{https://huggingface.co/collections/asdjghh/thinking-reward-model}}}

\begin{document}
\maketitle

\input{Sec/intro}
\input{Sec/related}
\input{Sec/pre}
\input{Sec/exp}
\input{Sec/conclusion}

\newpage

\bibliography{ref}
\bibliographystyle{plainnat}


\newpage
\appendix
\onecolumn

\input{Sec/appendix}

\end{document}

%% file: Sec/intro.tex

\begin{figure}[htbp]
    \centering
    \includegraphics[width=\textwidth]{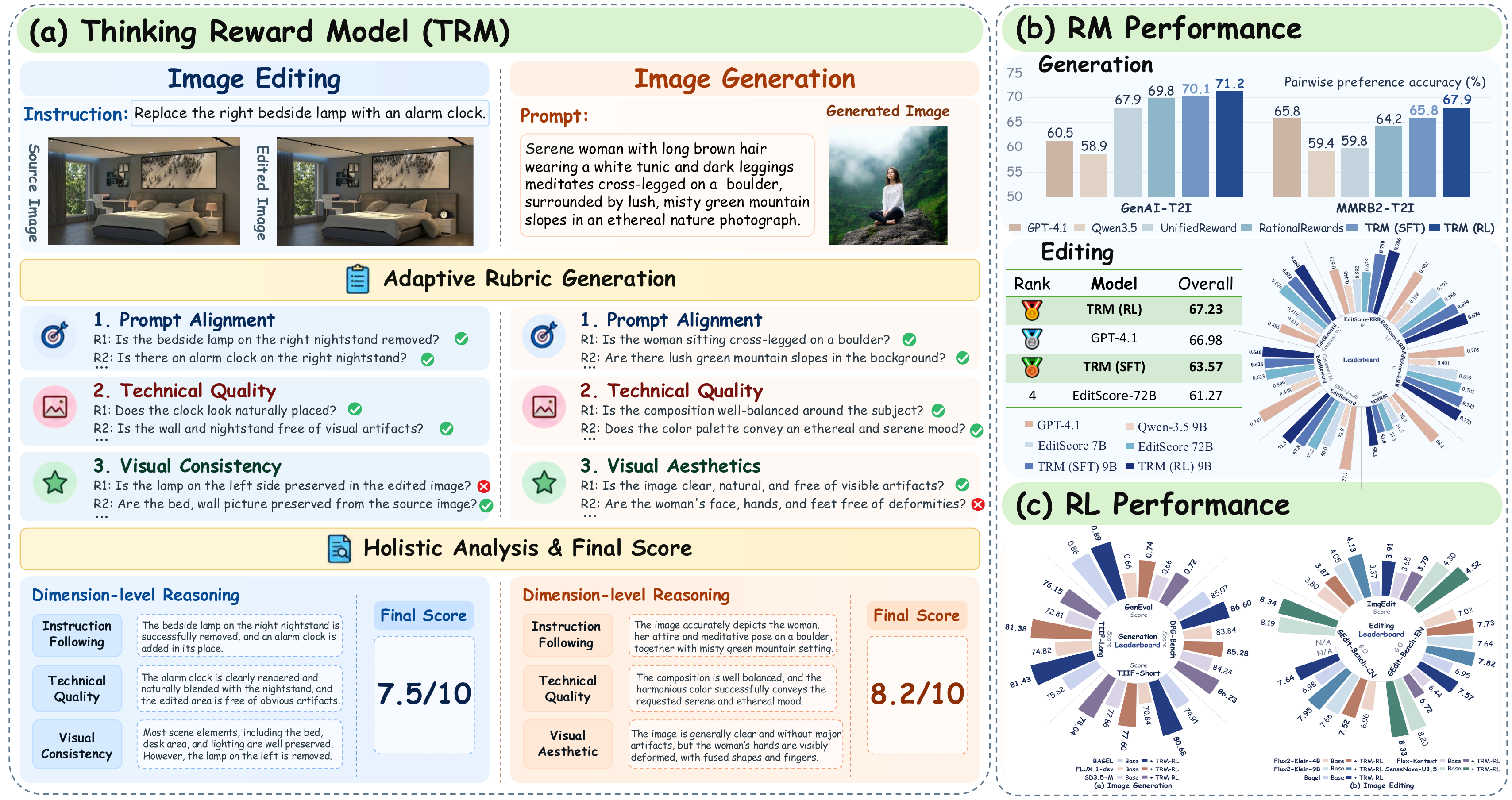}
    \caption{\textbf{Thinking Reward Model (\name) follows the ``Think Before You Score'' paradigm.}
    Given a visual generation case, \name first formulates case-adaptive rubrics that specify \emph{what matters}, then performs rubric-guided assessment before producing the final pointwise reward. This unified process achieves strong reward-modeling performance across image generation and editing, and effectively guides reinforcement learning for diverse visual generation models.}
    \label{fig:overview}
    \vspace{-0.5cm}
\end{figure}

\section{Introduction}
Scaling data and models~\citep{chen2025opengpt,wang2025gpt} has driven rapid advances in visual generation~\citep{wu2025qwen,diao2026sensenova,wang2026unireason}, particularly in image generation and editing. 
Yet models trained primarily with pretraining and supervised fine-tuning can still fall short of human expectations, as standard denoising and flow-matching objectives model data distributions without explicitly capturing preferences for instruction adherence, visual consistency, and perceptual quality. 
To better align generation with human preferences, recent work increasingly adopts reinforcement learning (RL) with preference-based reward signals~\citep{liu2026flow,xu2026qwen,guo2026leveraging}. 
Reward models are therefore central to RL-based post-training, translating human preferences into optimization signals for generative models.

Existing visual reward models~\citep{xu2023imagereward,wu2025editreward} typically map task conditions and candidate outputs to scalar scores or preference judgments. While capturing human preferences, direct scoring leaves the evaluation process largely implicit, obscuring whether specific requirements are satisfied or violated. Recent approaches~\cite{luo2026editscore,wang2026rationalrewards,wang2025unified} introduce reasoning prior to scoring, providing more explicit rationales for their assessments. However, visual evaluation is inherently multifaceted and instance-dependent: different examples may call for different evaluation criteria and exhibit distinct failure modes~\citep{han2024evalmuse, tang2026rewardverserubricguidedpolicyoptimization, tang2026evolutionaryagenticapproachopenended}. Consequently, even reasoning-based scoring leaves a fundamental question underexplored: \emph{what should be evaluated for this particular case?}

Our key observation is that evaluation criteria vary across tasks and individual cases, but the process of deriving and applying them can be shared across tasks. Inspired by how people make task-specific judgments, an evaluator first understands the task requirements and identifies the criteria
relevant to the current case, then examines
the candidate against each criterion and integrates the resulting evidence into an overall judgment. Effective evaluation thus first determines \emph{what matters} before judging \emph{how well the candidate performs}. We refer to this principle as \textbf{``Think Before You Score''}.

Following this principle, we propose the
\textbf{T}hinking \textbf{R}eward \textbf{M}odel (\textbf{\name{}}).
As illustrated in Figure~\ref{fig:overview}, \name{} first
generates a case-adaptive rubric that specifies the evaluation
criteria for the current task and candidate.
It then assesses the candidate against each criterion,
integrates the resulting evidence into a holistic judgment,
and outputs a pointwise reward.
By making the evaluation criteria explicit, the rubric
connects task interpretation with quality assessment,
transforming an implicit condition-to-score mapping into
a structured, case-adaptive evaluation process.

Building an effective thinking reward model requires both learning a structured evaluation process and capturing fine-grained quality differences.
To train such a model,  we construct diverse training data for image generation and image editing through a unified pipeline spanning multiple tasks and difficulty levels, together with a two-stage human–AI annotation process that provides high-quality rubrics and scores. 
Supervised fine-tuning on these data establishes the rubric-guided evaluation capability, while pairwise preference supervision further enhances fine-grained reward discrimination. 
A natural approach is to incorporate this preference supervision
through a Bradley--Terry-style objective applied directly to
pointwise scores. However, this objective continues to encourage
larger score margins even after the preference ordering is
correct, which can lead to increasingly polarized score
distributions.
We therefore introduce \textbf{Pairwise Dual-Group Relative Policy Optimization (\PDGRPO)}, which leverages pairwise preferences through response-level relative optimization while retaining fine-grained pointwise scoring. 
Extensive experiments across image generation and image editing benchmarks demonstrate the effectiveness of \name, while its application to generation model optimization further shows that stronger reward modeling translates into improved generation quality.

Our main contributions are summarized as follows:

\begin{itemize}[leftmargin=*]
    \item We introduce \textbf{Think Before You Score}, a visual reward modeling paradigm that explicitly determines \emph{what to evaluate} for each case before deciding \emph{how to score}.

    \item We construct a diverse dataset of approximately 20K image generation examples and 28K image editing examples through a unified pipeline with two-stage human-AI annotation, providing structured supervision with case-adaptive rubrics, criterion-level assessments, and quality scores.

    \item We develop \name through cold-start SFT followed by \textbf{\PDGRPO}, which leverages pairwise preferences to improve fine-grained pointwise discrimination while mitigating score polarization.

    \item Extensive experiments demonstrate that \name{} achieves strong performance on image generation and editing reward-modeling benchmarks and provides effective reward signals for improving diverse visual generation models through reinforcement learning.
\end{itemize} 

%% file: Sec/related.tex
\section{Related Work}

\textbf{Reward Models for Visual Generation.}
Visual reward modeling has gained increasing attention with advances in visual generation. 
Existing methods mainly follow regressive or generative paradigms: regressive approaches predict scalar rewards from task conditions and candidate outputs~\citep{xu2023imagereward,wu2025editreward}, while generative approaches leverage multimodal models to produce quality assessments, increasingly with explicit analysis or reasoning before scoring~\citep{luo2026editscore,wang2026rationalrewards}. 
Evaluation can be pointwise or pairwise, with image generation typically focusing on prompt adherence and visual quality~\citep{huang2026alphagrpo}, and image editing additionally considering edit correctness and content preservation~\citep{bai2026edit, zhang2026well}. 
As summarized in Table~\ref{tab:reward_model_comparison}, existing methods largely rely on fixed evaluation criteria, despite substantial variation in requirements and potential failure modes across individual cases.

\begin{table*}[htbp]
\centering
\caption{Comparison of representative reward models.
}
\label{tab:reward_model_comparison}
\vspace{-7pt}
\setlength{\tabcolsep}{6pt}
\renewcommand{\arraystretch}{1.12}
\resizebox{0.96\textwidth}{!}{%
\begin{tabular}{l|c|c|cc|ccc}
\toprule
\multirow[c]{2}{*}[-2pt]{\textbf{Method}} &
\multirow[c]{2}{*}[-2pt]{\textbf{Task}} &
\multirow[c]{2}{*}[-2pt]{\textbf{Modeling Paradigm}} &
\multicolumn{2}{c|}{\textbf{Scoring}} &
\multirow[c]{2}{*}[-2pt]{%
\shortstack[c]{\textbf{Adaptive}\\\textbf{Rubrics}}} &
\multirow[c]{2}{*}[-2pt]{%
\shortstack[c]{\textbf{Fine-Grained}\\\textbf{Verification}}} &
\multirow[c]{2}{*}[-2pt]{%
\shortstack[c]{\textbf{RL}\\\textbf{Optimization}}} \\
\cmidrule(lr){4-5}
& & & \textbf{Point} & \textbf{Pair} & & & \\
\midrule
ImageReward
& T2I & Regressive
& \cmark & --
& \xmark & \xmark & \xmark \\

UnifiedReward
& T2I, T2V & Generative
& \cmark & \cmark
& \xmark & \cmark & \xmark \\


RationalRewards
& TI2I, T2I & Generative
& \cmark & \cmark
& \xmark & \cmark & \xmark \\

RewardDance
& T2I & Generative
& -- & \cmark
& \xmark & \cmark & \xmark \\

FIRM-Reward
& TI2I, T2I & Generative
& \cmark & --
& \xmark & \cmark & \xmark \\

EditReward
& TI2I & Regressive
& \cmark & --
& \xmark & \xmark & \xmark \\

EditScore
& TI2I & Generative
& \cmark & --
& \xmark & \cmark & \xmark \\

\midrule
\textbf{\name{} (Ours)}
& \textbf{TI2I, T2I} & \textbf{Generative}
& \cmark & --
& \cmark & \cmark & \cmark \\
\bottomrule
\end{tabular}%
}
\end{table*}

\textbf{Reinforcement Learning for Visual Generation.}
Reward models play an important role in aligning visual
generation with human preferences through preference
optimization and reinforcement
learning~\citep{luo2026editscore,wang2026rationalrewards}.
Early studies explored diffusion policy optimization and
reward-based fine-tuning~\citep{ren2025diffusion,xu2023imagereward}.
Flow-GRPO~\citep{liu2026flow} extends online reinforcement learning to
flow-matching models by converting deterministic ODE
sampling into stochastic SDE sampling and optimizing
policies with group-relative advantages.
Building on this framework, subsequent studies improve
sampling efficiency and quality~\citep{wang2025coefficients},
introduce policy update constraints, and refine preference-based
advantage estimation~\citep{yang2026flowguard}
to use reward feedback more efficiently and stably for
generation model optimization.
Complementary to these optimization methods, our work
focuses on the quality of reward feedback itself.
We use pairwise preference supervision to strengthen
fine-grained pointwise reward modeling and apply the
resulting rewards to reinforcement learning for image
generation and editing.

%% file: Sec/pre.tex

\section{A Unified Reward Modeling Paradigm for Visual Generation}

\subsection{Can Reward Modeling Be Unified Across Visual Generation Tasks?}

Visual generation encompasses tasks with different input conditions and evaluation requirements, yet their reward modeling shares a common goal: estimating how well a candidate output satisfies the given task condition. 
We study image generation and image editing as two representative tasks and abstract each case as $x=(c,o)$, where $c$ denotes the task condition and $o$ the candidate output. 
For image generation, $c$ is a text prompt; for image editing, it consists of a source image and an editing instruction.
A general reward model can then be formulated as $f_{\theta}(sp,c,o)\rightarrow r$, where $sp$ specifies the evaluation protocol and $r$ is the predicted pointwise reward. 
Although concrete conditions and criteria vary across tasks, the underlying evaluation process is shared: an evaluator understands the task requirements, determines what to check, inspects the candidate accordingly, and aggregates the observations into a final reward. 
This suggests a unified paradigm that shares the evaluation process across tasks while adapting the concrete criteria to each case.
However, a critical question remains: \textbf{how should the model determine what to check for each individual case before assigning a score?}

\subsection{How Should a General Reward Model Evaluate a Visual Generation Case?}
\label{rubric}

We observe that what should be checked varies across cases, even within the same task. 
The task condition specifies the requirements, while the candidate output may introduce aspects or potential issues, such as object relations and visual defects in image generation, or identity preservation and unintended changes in image editing. 
Evaluation criteria should therefore adapt to both the task requirements and candidate output. 
However, conventional visual reward models typically map them directly to a scalar reward, leaving such case-adaptive criteria implicit. 
Consequently, case-specific requirements and fine-grained issues may not be adequately reflected in the final reward.

We therefore argue that a reward model should explicitly determine \textbf{what to check} before deciding \textbf{how to score}, a paradigm we term \textbf{Think Before You Score}. 
Specifically, a Thinking Reward Model (TRM) structures evaluation as $R\rightarrow J\rightarrow H\rightarrow s$, where $R$ denotes the case-adaptive rubric, $J$ the rubric-level inspections and judgments, $H$ the dimension-level assessments summarizing these judgments, and $s$ the final pointwise reward. 
This forms a determine–inspect–aggregate–score process. Crucially, $R$ is not a fixed checklist but an evaluation plan instantiated from the task condition and candidate output before the corresponding judgments and final reward are formed. 
This enables a consistent evaluation procedure with adaptive case-level criteria.

We instantiate this process with three high-level dimensions. Both image generation and editing share \textit{Prompt Alignment} and \textit{Visual Quality}, while the third dimension is task-specific: \textit{Aesthetics} for image generation and \textit{Source Consistency} for image editing. Within each dimension, TRM generates atomic, case-adaptive rubrics and inspects the candidate to produce a binary Yes/No judgment for each. 
These judgments constitute $J$ and are summarized into dimension-level assessments $H$, from which TRM predicts the final pointwise reward $s$. 
Thus, the high-level dimensions provide a consistent evaluation structure, while the concrete rubrics adapt to each case.

\begin{figure}[t]
\centering
\includegraphics[width=\linewidth]{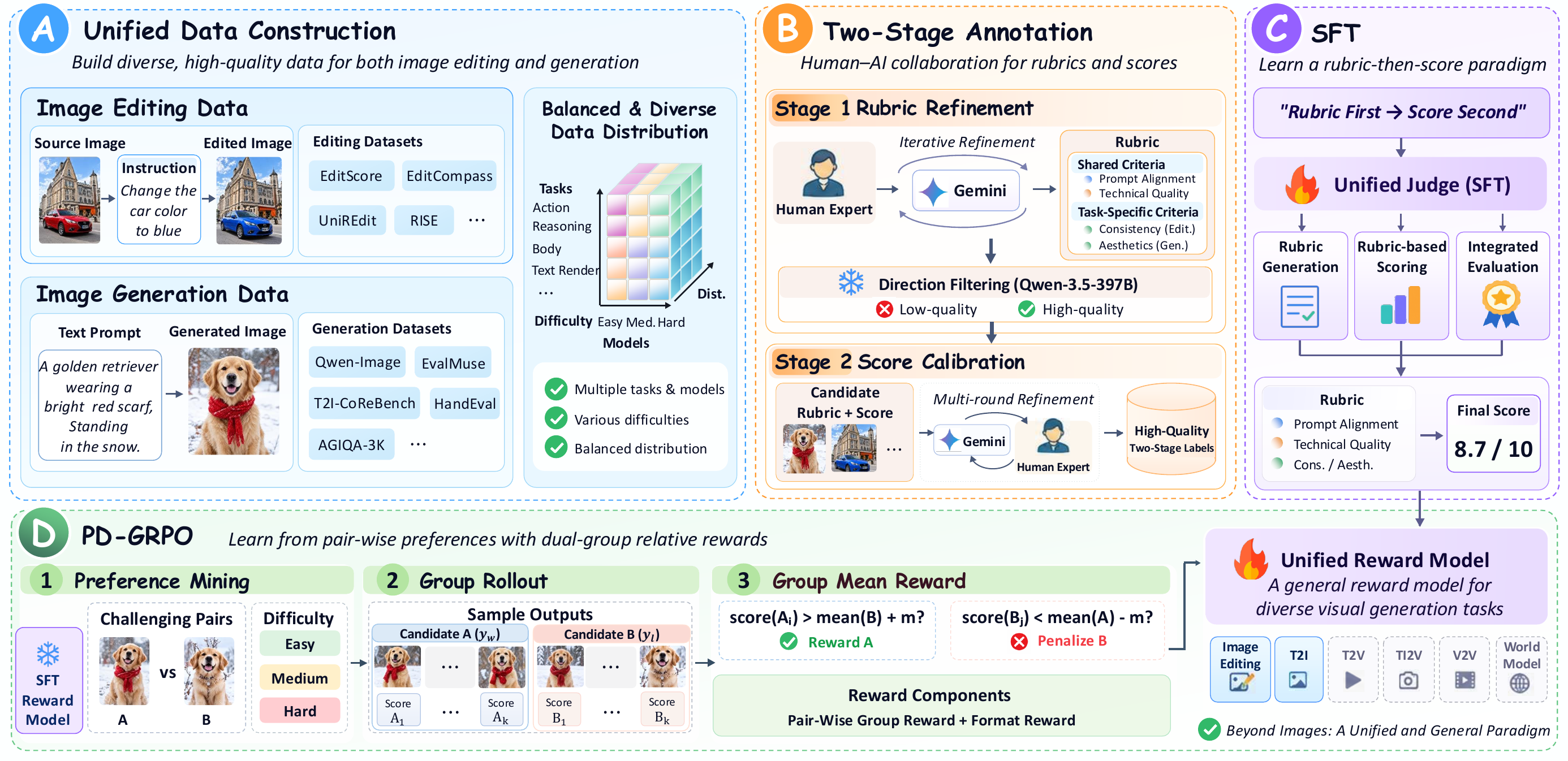}
\caption{\textbf{Overview of data construction and training pipeline.} We construct diverse and balanced data for image generation and editing, annotate rubrics and scores through a two-stage process, learn rubric-then-score evaluation via SFT, and further improve pointwise scoring with \PDGRPO using difficulty-aware pairwise preference supervision.}
\label{fig:pipeline}
\vspace{-0.5cm}
\end{figure}

\section{Thinking Reward Model}
\label{sft}

\subsection{Data Construction and Cold-Start SFT}

As shown in Figure~\ref{fig:pipeline}, training the Thinking Reward Model requires not only accurate scores, but also diverse visual cases and high-quality structured evaluation traces. 
Despite their different inputs and evaluation requirements, we adopt a unified pipeline for image generation and editing: constructing diverse cases, annotating rubrics and scores, and performing cold-start SFT to learn rubric-then-score evaluation.
Based on the resulting SFT model, we further construct difficulty-aware preference pairs for subsequent reinforcement learning (RL).

\textbf{Step 1: Raw Case Construction.}
Following prior work~\cite{luo2026editscore,wu2025editreward}, we curate existing data to obtain reliable evaluation samples. 
For image editing, we filter instruction–source image pairs for compatibility, while for image generation, we remove invalid or unreliable text–image cases. 
We further broaden task coverage with representative benchmarks, such as Edit-Compass~\citep{bai2026edit} and UniREdit-Bench~\citep{han2026unireditbench} for image editing, and Qwen-Image-Bench~\citep{li2026qwen} and EvalMuse~\citep{han2024evalmuse} for image generation.
We further perform rollouts with open-source and proprietary models of varying capabilities to increase output diversity. 
The resulting cases cover diverse task categories, model sources, quality levels, and failure modes, and are balanced across tasks, sources, and difficulty levels.

\textbf{Step 2: Expert-in-the-Loop Structured Annotation.}
We construct structured supervision through a two-stage human–AI annotation process. 
In the first stage, human experts iteratively refine the system prompt, while a teacher model produces case-adaptive rubrics that are reviewed for coverage, atomicity, redundancy, and verifiability. 
In the second stage, the verified rubrics guide the teacher model to produce rubric-level judgments and final scores, followed by expert calibration. 
The two stages respectively establish \textbf{what should be evaluated} and \textbf{how it should be evaluated}, yielding reliable structured evaluation traces for training. Through this process, we construct a high-quality dataset comprising
approximately 20K image generation cases and 28K image editing cases.

\textbf{Step 3: Cold-Start SFT.}
Using these structured annotations, we perform SFT to learn the rubric-then-score evaluation process. 
We construct three complementary training formats: \textit{Rubric Generation} for learning what to check, \textit{Rubric-based Scoring} for evaluating candidates against given rubrics, and \textit{Integrated Evaluation} for learning the complete structured evaluation. 
We jointly train all three formats in a single SFT stage to learn the complete $R \rightarrow J \rightarrow H \rightarrow s$ process. 
The resulting model, denoted as \textbf{\name (SFT)}, serves as the initialization for subsequent RL.

\textbf{Step 4: Difficulty-Aware Preference Pair Construction.}
Starting from \name (SFT), we construct preference pairs from candidate outputs under the same task condition. 
We rank each pair by their rewards and use the reward gap as a proxy for preference difficulty: larger gaps indicate easier comparisons, whereas smaller gaps require finer-grained discrimination. 
We partition the pairs into easy, medium, and hard subsets and sample across all three levels to cover both clear preferences and subtle quality differences. 
Finally, we balance the data across task categories, candidate sources, and difficulty levels, yielding approximately $4,000$ high-quality preference pairs.

\subsection{Pairwise Preference Optimization}
\label{optimization}
Although \name(SFT) learns the complete pointwise evaluation process, pointwise supervision treats each candidate independently and does not explicitly exploit relative preferences between candidates. 
This distinction becomes particularly important when candidates have similar overall quality but differ in subtle yet meaningful aspects. 
In image editing, for instance, two candidates may both satisfy the editing instruction and receive similarly high scores, while differing in realism or integration with the surrounding scene. 
Such fine-grained differences can still induce clear relative preferences. 
We therefore introduce pairwise preference supervision to improve fine-grained discrimination between candidates while retaining the original pointwise reward interface.

As illustrated in Figure~\ref{fig:pipeline}, for each preference pair, we independently sample multiple pointwise evaluation responses for the preferred and dispreferred candidates and use their relative scores to derive the training signal. 
During reinforcement learning, the reward is defined as $R(y)=r_{\mathrm{pref}}(y)+\lambda_{\mathrm{fmt}}r_{\mathrm{fmt}}(y)$, where $r_{\mathrm{pref}}$ translates the pairwise preference into an optimization signal for pointwise reward prediction, while $r_{\mathrm{fmt}}$ encourages valid structured outputs. 
Since $r_{\mathrm{fmt}}$ remains fixed throughout training, we focus on the design of $r_{\mathrm{pref}}$ below.

\textbf{Bradley–Terry as a Natural Starting Point.}
The Bradley–Terry (BT) model~\citep{bradley1952rank} provides a natural way to incorporate pairwise supervision while preserving the pointwise scoring interface. 
Given a rollout $i$ from the preferred candidate with score $s_i^{+}$ and a rollout $j$ from the dispreferred candidate with score $s_j^{-}$, we define $M_{ij}=\sigma\left((s_i^{+}-s_j^{-}-m)/\tau\right)$, where $M_{ij}$ is the probability that $i$ is preferred over $j$, $m$ is the preference margin, and $\tau$ is the temperature.

In initial BT-style formulation, each rollout is rewarded by its average pairwise preference against rollouts from the opposite side, defined as $r_i^{+}=\frac{1}{|G^{-}|}\sum_{j\in G^{-}}M_{ij}$ and $r_j^{-}=\frac{1}{|G^{+}|}\sum_{i\in G^{+}}M_{ij}$, where $G^{+}$ and $G^{-}$ denote the preferred and dispreferred rollout sets, respectively.

In practice, as shown in Figure~\ref{fig:compare_score}, directly optimizing this reward leads to pronounced \textbf{score polarization}: preferred scores progressively increase while dispreferred scores decrease. 
This follows from the monotonicity of $M_{ij}$ with respect to $s_i^{+}-s_j^{-}$: enlarging the score gap always increases the preference reward, even when the pair is already correctly ordered. 
Thus, BT enforces relative ordering without constraining the absolute pointwise scores.

Consequently, the objective has no interior optimum with respect to the score gap. 
In a bounded scoring space, continued optimization drives the two sides toward opposite boundaries, yielding polarized rather than well-calibrated scores. 
This motivates an objective that enforces sufficient relative separation but stops rewarding further gap expansion once that separation is achieved.

\textbf{Pairwise Dual-Group Relative Policy Optimization.}
Motivated by the above analysis, we propose Pairwise Dual-Group Relative Policy Optimization (\PDGRPO{}), which incorporates pairwise preference supervision while assigning credit to independently generated pointwise evaluations.

Given a preference pair $(x^{+},x^{-})$, we independently sample $N$ pointwise responses for each candidate, forming $G^{+}={y_1^{+},\ldots,y_N^{+}}$ and $G^{-}={y_1^{-},\ldots,y_N^{-}}$ for the preferred and dispreferred candidates, respectively. 
Each rollout independently produces a structured evaluation and pointwise score without observing the other candidate.

We compute the group means as $\mu^{+}=\frac{1}{N}\sum_{i=1}^{N}s_i^{+}$ and $\mu^{-}=\frac{1}{N}\sum_{j=1}^{N}s_j^{-}$. 
The two groups serve as mutual references, with each rollout evaluated against the mean score of the opposite group, as illustrated in Figure~\ref{fig:pipeline}. 
Specifically, $r_{\mathrm{pref}}(y)=\mathbb{1}[s(y)-\mu^{-}>m]$ for $y\in G^{+}$ and $r_{\mathrm{pref}}(y)=\mathbb{1}[\mu^{+}-s(y)>m]$ for $y\in G^{-}$, where $m$ is the required separation margin. 
Thus, pairwise preferences provide response-level credit based on whether each pointwise score achieves sufficient separation from the opposite group.

Unlike the BT-style objective, this reward becomes constant once the required margin is satisfied, so further enlarging the score gap provides no additional benefit. 
\PDGRPO therefore enforces the desired relative separation without a persistent incentive toward score polarization. 
The margin $m$ can further be adjusted according to the difficulty of each preference pair.

The dual-group structure is used only for reward construction. For policy optimization, we combine all $2N$ responses into $\mathcal{G}=G^{+}\cup G^{-}$ and compute the group-relative advantage as $A_k=(R_k-\mu_{\mathcal{G}})/(\sigma_{\mathcal{G}}+\epsilon)$, where $\mu_{\mathcal{G}}$ and $\sigma_{\mathcal{G}}$ are the reward mean and standard deviation within $\mathcal{G}$. 
We then apply the standard clipped group-relative objective with KL regularization~\citep{shao2024deepseekmath}.

Since the two candidates interact only during reward construction, each remains independently evaluated at inference time. 
Thus, \PDGRPO exploits fine-grained pairwise supervision while preserving the pointwise inference interface of \name.

%% file: Sec/exp.tex

\section{Experiments}
\subsection{Experimental Setups}
We evaluate \name from two complementary perspectives across image generation and editing: reward-modeling performance on established benchmarks and effectiveness in guiding downstream reinforcement learning, assessed through both quantitative and qualitative results.


\textbf{Reward Modeling Benchmarks and Baselines.}
We evaluate \name{} on GenAI-T2I~\citep{li2024genai} and MMRB2-T2I~\citep{hu2026multimodal} for image generation, and EditScore-ERB~\citep{luo2026editscore}, MMRB2~\citep{hu2026multimodal}, EditReward-ERB~\citep{wu2025editreward}, and EditReward-Compass~\citep{bai2026edit} for image editing. 
We compare against proprietary multimodal models from the
GPT~\citep{gpt41} and Gemini~\citep{google2025gemini3pro} families, open-source models from the
Qwen~\citep{Qwen2.5-VL,Qwen3-VL,qwen3.5} family, and specialized reward models, including
HPSv3~\citep{ma2025hpsv3},
UnifiedReward~\citep{wang2025unified},
RationalRewards~\citep{wang2026rationalrewards},
EditScore~\citep{luo2026editscore}, and
FIRM-Reward~\citep{zhao2026trust}. 

\textbf{Visual Generation Benchmarks and Baselines.}
To evaluate \name{} as a training signal, we use it for
on-policy optimization of BAGEL~\citep{deng2025bagel}, FLUX.1-dev~\citep{flux2024}, and SD3.5-M
for image generation, and FLUX.2-Klein (4B/9B)~\citep{flux-2-2025}, BAGEL,
FLUX-Kontext, and SenseNova-U1.5~\citep{diao2026sensenova} for image editing.
We evaluate image generation on GenEval~\citep{ghosh2023geneval}, DPG-Bench~\citep{hu2024ella}, and TIIF-testmini~\citep{wei2025tiif}, and image editing on ImgEdit~\citep{ye2026imgedit} and GEdit-Bench (EN/CN)~\citep{liu2025step1x}. 

\textbf{Implementation Details.}
\name is initialized from Qwen3.5-9B~\citep{qwen3.5} and trained with cold-start SFT followed by \PDGRPO on approximately $4,000$ preference pairs. 
Both stages use LoRA, with $8$ rollouts per candidate. 
We set the separation margin to $m=0$ for image editing and $m=0.05$ for image generation, with scores normalized before reward computation. 
Qwen3.5-9B (Baseline) denotes the original model evaluated under the same pointwise protocol without reward-model training.
More details are provided in the Appendix~\ref{supp:training_details}.

\subsection{Reward Model Performance}



\begin{wraptable}{r}{0.40\textwidth}
\centering
\vspace{-0.7cm}
\caption{
Performance on image generation reward-modeling benchmarks.
}

\label{tab:t2i_reward_results}

\setlength{\tabcolsep}{5pt}
\renewcommand{\arraystretch}{1.12}

\resizebox{0.9\linewidth}{!}{%
\begin{tabular}{lccc}
\toprule
\textbf{Model} &
\textbf{Size} &
\textbf{GenAI-T2I} &
\textbf{MMRB2-T2I} \\
\midrule

\multicolumn{4}{>{\columncolor[HTML]{F0F5FF}}c}
{\textit{\textbf{Proprietary Models}}} \\
\textcolor{LightGray}{GPT-4.1} & -- & \textcolor{LightGray}{60.5} & \textcolor{LightGray}{65.8} \\
\textcolor{LightGray}{Gemini 2.5 Flash} & -- & \textcolor{LightGray}{65.8} & \textcolor{LightGray}{63.1} \\
\textcolor{LightGray}{Gemini 2.5 Pro} & -- & \textcolor{LightGray}{66.2} & \textcolor{LightGray}{70.5} \\
\textcolor{LightGray}{Gemini 3 Pro} & -- & \textcolor{LightGray}{73.1} & \textcolor{LightGray}{74.4} \\
\midrule

\multicolumn{4}{>{\columncolor[HTML]{FFF6EA}}c}
{\textit{\textbf{Open-Source Models}}} \\
Qwen2.5-VL & 7B & -- & 50.4 \\
Qwen2.5-VL & 72B & 66.6 & 59.1 \\
Qwen3-VL & 8B & 55.1 & 59.4 \\
Qwen3-VL & 32B & 66.9 & 64.1 \\
Qwen-3.5 & 9B & 66.1 & 64.5 \\
\midrule

\multicolumn{4}{>{\columncolor[HTML]{F7F7F7}}c}
{\textit{\textbf{Image Generation Reward Models}}} \\
HPSv3 & 7B & 70.8 & 60.2 \\
UnifiedReward & 7B & 67.9 & 59.8 \\
RationalRewards & 8B & 69.8 & 64.2 \\
\midrule

\multicolumn{4}{>{\columncolor[HTML]{F1FFF5}}c}
{\textit{\textbf{Our Models}}} \\
Qwen-3.5 (Baseline) & 9B & 58.9 & 59.4 \\
\name{} (SFT) & 9B & 70.1 & 65.8 \\
\textbf{\name{} (RL)} & 9B & \textbf{71.2} & \textbf{67.9} \\
\bottomrule
\end{tabular}%
}
\vspace{-1em}
\end{wraptable}

\textbf{Image Generation.}
As shown in Table~\ref{tab:t2i_reward_results}, \name{} achieves strong preference modeling across both benchmarks.
Cold-start SFT substantially improves the Qwen-3.5 baseline from 58.9\%/59.4\% to 70.1\%/65.8\% on GenAI-T2I/MMRB2-T2I, demonstrating the effectiveness of structured rubric-guided training.
\PDGRPO{} further improves the results to 71.2\%/67.9\%, with gains of 1.1 and 2.1 percentage points over SFT.
Notably, our 9B model outperforms all evaluated open-source models and specialized reward models on MMRB2-T2I, while also surpassing GPT-4.1 on both benchmarks.
These results show that structured SFT establishes strong pointwise evaluation, while pairwise preference optimization further improves fine-grained discrimination.
For pointwise models, we report accuracy over non-tied predictions; tie-aware results are provided in Appendix~\ref{sec:tie_analysis}.

\textbf{Image Editing.}
Table~\ref{tab:editing_rm_results} shows similarly consistent improvements on image editing. 
Compared with the Qwen-3.5 baseline, \name (SFT) substantially improves all reported metrics, reaching 0.750/0.639/0.743 on EditScore-ERB and 67.8\% on EditReward-ERB. 
\PDGRPO further improves every metric, including MMRB2 from 53.0\% to 58.2\% and EditReward-Compass to 0.640/0.660. 
Despite using only 9B parameters, \name (RL) also consistently outperforms the specialized 72B EditScore~\citep{luo2026editscore} model across all reported benchmarks and metrics. 
Together, these results demonstrate that pairwise preference optimization consistently strengthens the pointwise evaluation capability established by SFT across diverse editing criteria.

\begin{table*}[htbp]
\centering
\caption{
Performance on image editing reward-modeling benchmarks.
}

\label{tab:editing_rm_results}
\setlength{\tabcolsep}{5.5pt}
\renewcommand{\arraystretch}{1.12}

\resizebox{0.75\textwidth}{!}{
\begin{tabular}{lcccccccc}
\toprule
\textbf{Model} &
\textbf{Size} &
\multicolumn{3}{c}{\textbf{EditScore-ERB}} &
\textbf{MMRB2} &
\textbf{EditReward-ERB} &
\multicolumn{2}{c}{\textbf{EditReward-Compass}} \\
\cmidrule(lr){3-5}
\cmidrule(lr){8-9}
& &
\textbf{IF} &
\textbf{VC} &
\textbf{O} &
&
\textbf{2-path} &
\textbf{IA} &
\textbf{VC} \\
\midrule

\rowcolor[HTML]{F0F5FF}
\multicolumn{9}{c}{\textit{\textbf{Proprietary Models}}} \\
\textcolor{LightGray}{GPT-4.1} & -- & \textcolor{LightGray}{0.673} & \textcolor{LightGray}{0.602} & \textcolor{LightGray}{0.705} & \textcolor{LightGray}{68.2} & \textcolor{LightGray}{72.1} & \textcolor{LightGray}{0.747} & \textcolor{LightGray}{0.485} \\
\textcolor{LightGray}{GPT-5} & -- & \textcolor{LightGray}{0.777} & \textcolor{LightGray}{0.669} & \textcolor{LightGray}{0.755} & \textcolor{LightGray}{73.8} & \textcolor{LightGray}{73.0} & \textcolor{LightGray}{--} & \textcolor{LightGray}{--} \\
\textcolor{LightGray}{Gemini 2.5 Pro} & -- & \textcolor{LightGray}{0.703} & \textcolor{LightGray}{0.560} & \textcolor{LightGray}{0.722} & \textcolor{LightGray}{71.3} & \textcolor{LightGray}{78.3} & \textcolor{LightGray}{--} & \textcolor{LightGray}{--} \\
\textcolor{LightGray}{Gemini 3.1 Pro} & -- & \textcolor{LightGray}{0.877} & \textcolor{LightGray}{0.716} & \textcolor{LightGray}{0.841} & \textcolor{LightGray}{74.9} & \textcolor{LightGray}{73.9} & \textcolor{LightGray}{0.832} & \textcolor{LightGray}{0.600} \\

\midrule
\rowcolor[HTML]{FFF6EA}
\multicolumn{9}{c}{\textit{\textbf{Open-Source Models}}} \\
Qwen2.5-VL & 7B & 0.458 & 0.325 & 0.432 & 55.2 & 63.4 & 0.427 & 0.217 \\
Qwen2.5-VL & 32B & 0.498 & 0.376 & 0.563 & 67.3 & 65.2 & 0.612 & 0.412 \\
Qwen2.5-VL & 72B & 0.540 & 0.435 & 0.621 & 65.8 & 67.8 & 0.637 & 0.421 \\
Qwen3-VL & 8B & 0.383 & 0.239 & 0.571 & 62.0 & 60.9 & 0.565 & 0.365 \\
Qwen-3.5 & 9B & 0.612 & 0.389 & 0.500 & 51.4 & 47.0 & 0.585 & 0.452 \\

\midrule
\rowcolor[HTML]{F7F7F7}
\multicolumn{9}{c}{\textit{\textbf{Image Editing Reward Models}}} \\
EditScore & 7B
& 0.592 & 0.591 & 0.659
& 51.3
& 60.0
& 0.509 & 0.416 \\

EditScore & 72B
& 0.635 & 0.586 & 0.703
& 53.3
& 65.2
& 0.623 & 0.626 \\

FIRM-Reward & 8B
& 0.476 & 0.565 & 0.607
& 44.5
& 50.9
& 0.520 & 0.507 \\

\midrule
\rowcolor[HTML]{F1FFF5}
\multicolumn{9}{c}{\textit{\textbf{Our Models}}} \\
Qwen-3.5 (Baseline) & 9B
& 0.440 & 0.308 & 0.401
& 30.9
& 33.8
& 0.448 & 0.314 \\

\name{}(SFT) & 9B
& 0.750 & 0.639 & 0.743
& 53.0
& 67.8
& 0.626 & 0.622 \\

\textbf{\name{}(RL)} & 9B
& \textbf{0.786}
& \textbf{0.674}
& \textbf{0.773}
& \textbf{58.2}
& \textbf{71.3}
& \textbf{0.640}
& \textbf{0.660} \\

\bottomrule
\end{tabular}
}
\end{table*}

\subsection{Reward-Guided RL Optimization}
\label{RL}
Reward-modeling benchmarks measure the evaluation capability of \name, but a practical reward model should also provide effective training signals for improving visual generation. 
We therefore use \name to guide FlowGRPO~\citep{liu2026flow} optimization of image generation and editing models, evaluating whether its learned rewards translate into consistent gains in generation quality.

\begin{wraptable}{r}{0.6\textwidth}
\centering
\caption{
Results of \name{}-guided RL on image generation.
}
\label{tab:image_generation_rl}
\small
\setlength{\tabcolsep}{5pt}
\renewcommand{\arraystretch}{1.08}
\resizebox{\linewidth}{!}{%
\begin{tabular}{lcccc}
\toprule
\textbf{Model}
& \textbf{GenEval}$\uparrow$
& \textbf{DPG-Bench}$\uparrow$
& \textbf{TIIF-Short}$\uparrow$
& \textbf{TIIF-Long}$\uparrow$ \\
\midrule

\multicolumn{5}{l}{\textit{\textbf{Representative Image Generation Models}}} \\
OmniGen2
    & 0.80 & 83.60 & 70.20 & 70.30 \\
LongCat-Image
    & 0.87 & 86.80 & -- & -- \\
Qwen-Image
    & 0.87 & 88.32 & 86.14 & 86.83 \\
LLaDA-Image
    & 0.85 & 87.48 & -- & -- \\
Z-Image
    & 0.84 & 88.14 & 80.20 & 83.01 \\

\midrule
\multicolumn{5}{l}{\textit{\textbf{\name{}-guided Reinforcement Learning}}} \\

BAGEL
    & 0.86 & 85.07 & 74.91 & 75.62 \\
\quad + \name{}-guided RL
    & \textbf{0.89}
    & \textbf{86.60}
    & \textbf{80.68}
    & \textbf{81.43} \\

\addlinespace[2pt]

FLUX.1-dev
    & 0.66 & 83.84 & 70.84 & 74.82 \\
\quad + \name{}-guided RL
    & \textbf{0.73}
    & \textbf{85.28}
    & \textbf{77.60} & \textbf{81.38} \\

\addlinespace[2pt]

SD3.5-M
    & 0.66 & 84.24 & 72.86 & 72.81 \\
\quad + \name{}-guided RL
    & \textbf{0.72} & \textbf{86.23} & \textbf{78.04} & \textbf{76.15} \\

\bottomrule
\vspace{-0.5em}
\end{tabular}%
}
\end{wraptable}

\textbf{Image Generation.}
As shown in Table~\ref{tab:image_generation_rl}, \name-guided reinforcement learning consistently improves BAGEL, FLUX.1-dev, and SD3.5-M across all evaluated benchmarks, demonstrating that the learned reward generalizes across model families and capability levels.
The improvements cover GenEval, DPG-Bench, and both TIIF subsets, with particularly notable gains on fine-grained instruction following.
Specifically, BAGEL improves by 5.81 points on TIIF-Long, while FLUX.1-dev gains 6.76 and 6.56 points on TIIF-Short and TIIF-Long, respectively.

The qualitative comparisons in Figures~\ref{fig:qual_gen_bagel} and~\ref{fig:qual_gen_flux} further reveal where these gains arise.
After \name{}-guided optimization, both models better satisfy fine-grained compositional constraints, including object counting, composition, spatial relations, and text rendering.
BAGEL more accurately follows specified object counts and textual requirements, while FLUX.1-dev better preserves required objects in multi-constraint prompts and renders requested text more faithfully.
Together, these results show that \name{} provides effective training signals for improving fine-grained prompt alignment across diverse text-to-image models.

\begin{wraptable}{r}{0.57\textwidth}
\centering
\vspace{-0.5cm}
\caption{
Results of \name{}-guided RL on image editing.
}
\label{tab:image_editing_rl}
\setlength{\tabcolsep}{4pt}
\renewcommand{\arraystretch}{1.08}
\resizebox{\linewidth}{!}{%
\begin{tabular}{lccccccc}
\toprule
\textbf{Model} &
\textbf{ImgEdit} &
\multicolumn{3}{c}{\textbf{GEdit-Bench-EN}} &
\multicolumn{3}{c}{\textbf{GEdit-Bench-CN}} \\
\cmidrule(lr){3-5}
\cmidrule(lr){6-8}
& &
\textbf{G\_SC} &
\textbf{G\_PQ} &
\textbf{G\_O}$\uparrow$ &
\textbf{G\_SC} &
\textbf{G\_PQ} &
\textbf{G\_O}$\uparrow$ \\
\midrule

\multicolumn{8}{l}{\textit{\textbf{Representative Image Editing Models}}} \\
OmniGen2
& 3.44 & 7.16 & 6.77 & 6.41 & -- & -- & -- \\
LongCat-Image-Edit
& 4.44 & 8.13 & 8.18 & 7.75 & 8.14 & 8.12 & 7.73 \\
Qwen-Image-Edit2509
& 4.34 & 7.97 & 7.71 & 7.48 & 7.99 & 7.68 & 7.47 \\
LLaDA-Image
& -- & 8.04 & 7.18 & 7.34 & 7.71 & 7.59 & 7.29 \\
JoyAI-Image-Edit
& 4.46 & 8.83 & 8.12 & 8.28 & 8.62 & 8.11 & 8.13 \\
DeepGen1.0
& 4.14 & 7.66 & 7.12 & 7.17 & 7.22 & 7.31 & 6.82 \\

\midrule
\multicolumn{8}{l}{\textit{\textbf{\name{}-guided Reinforcement Learning}}} \\

Flux2-Klein-4B
& 3.80 & 7.68 & 7.31 & 7.02
& 7.65 & 7.27 & 6.96 \\

\quad + \name{}-guided RL
& \textbf{3.87}
& \textbf{8.20}
& \textbf{7.74}
& \textbf{7.73}
& \textbf{7.90}
& \textbf{7.65}
& \textbf{7.52} \\

Flux2-Klein-9B
& 4.05 & 8.39 & 7.57 & 7.64
& 8.37 & 7.59 & 7.66 \\

\quad + \name{}-guided RL
& \textbf{4.13}
& \textbf{8.43}
& \textbf{7.82}
& \textbf{7.82}
& \textbf{8.45}
& \textbf{7.95}
& \textbf{7.95} \\

BAGEL
& 3.37 & 7.89 & 6.68 & 6.95
& 7.92 & 6.76 & 6.98 \\

\quad + \name{}-guided RL
& \textbf{3.91}
& \textbf{8.39}
& \textbf{7.26}
& \textbf{7.57}
& \textbf{8.40}
& \textbf{7.33}
& \textbf{7.64} \\

Flux-Kontext
& 3.65 & 7.07 & 7.30 & 6.44
& -- & -- & -- \\

\quad + \name{}-guided RL
& \textbf{3.79}
& \textbf{7.22}
& \textbf{7.51}
& \textbf{6.72}
& -- & -- & -- \\

SenseNova-U1.5$^{\dagger}$
& 4.30 & 9.10 & 7.63 & 8.20
& \textbf{9.10} & 7.61 & 8.19 \\

\quad + \name{}-guided RL
& \textbf{4.52}
& \textbf{9.14}
& \textbf{7.83}
& \textbf{8.33}
& 9.09
& \textbf{7.87}
& \textbf{8.34} \\

\bottomrule
\end{tabular}%
}
\end{wraptable}

\textbf{Image Editing.}
As shown in Table~\ref{tab:image_editing_rl}, \name{}-guided reinforcement learning consistently improves diverse image-editing models across nearly all evaluated metrics, spanning Flux2-Klein at 4B and 9B scales, BAGEL, Flux-Kontext, and SenseNova-U1.5.
These consistent improvements across model families, scales, and initial capability levels demonstrate that \name{} provides effective training signals even for already strong image-editing models.
The gains are particularly pronounced for BAGEL, whose ImgEdit score increases from 3.37 to 3.91, while its overall scores on GEdit-Bench-EN/CN improve from 6.95/6.98 to 7.57/7.64.
Notably, the strong SenseNova-U1.5 baseline also benefits from \name{}-guided optimization, improving from 4.30 to 4.52 on ImgEdit and from 8.20/8.19 to 8.33/8.34 on GEdit-Bench-EN/CN.

\begin{figure}[t]
    \centering
    \resizebox{\linewidth}{!}{%
        \includegraphics{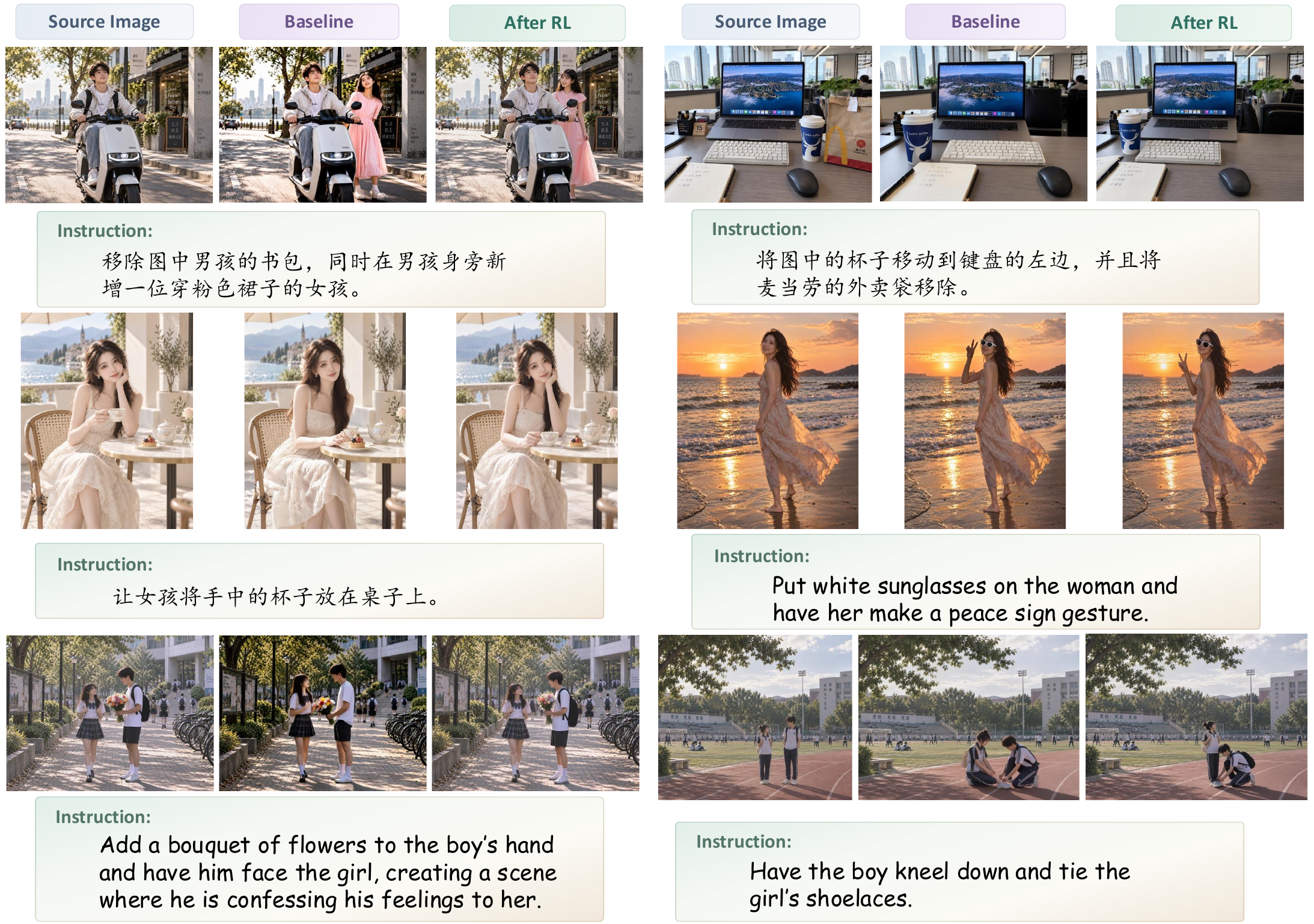} 
    }
    \caption{Qualitative comparison of SenseNova-U1.5 before and after
RL fine-tuning with \name.}
    \label{fig:show_case}
\end{figure}

Figure~\ref{fig:show_case} further illustrates the improvements under both Chinese and English editing instructions.
Compared with the baseline, \name{}-guided optimization better preserves unedited content while reducing unintended appearance changes and structural artifacts, and more faithfully follows the requested modifications.
These qualitative results complement the benchmark gains, demonstrating improved instruction following, content preservation, and visual quality across diverse editing scenarios.
Additional cases are provided in Figures~\ref{fig:qual_edit_bagel} and~\ref{fig:qual_edit_sensenova}.

\subsection{Ablation Studies}
\label{sec:ablation}

\paragraph{Image Generation.}
We compare \name with AlphaGRPO~\citep{huang2026alphagrpo} using the same BAGEL backbone. 
As shown in Table~\ref{tab:ablation_reward}, \name{} consistently outperforms AlphaGRPO across all evaluated benchmarks, with gains of 2.98 and 3.33 points on TIIF-Short and TIIF-Long, respectively. 
Improvements on GenEval and DPG-Bench further show that the gains generalize across complementary evaluation benchmarks. 
Under this controlled setting, these results demonstrate the effectiveness of \name{} as a reward signal for image generation optimization.

\begin{wraptable}{r}{0.53\textwidth}
\centering
\vspace{-0.5cm}
\caption{
Ablation and controlled comparisons of reward-guided optimization.
}
\label{tab:ablation_reward}

\scriptsize
\setlength{\tabcolsep}{3.2pt}
\renewcommand{\arraystretch}{1.05}

\resizebox{\linewidth}{!}{%
\begin{tabular}{lcccc}
\toprule
\multicolumn{5}{c}{\textbf{Image Generation: BAGEL}} \\
\midrule
Reward / Method
& GenEval$\uparrow$
& DPG-Bench$\uparrow$
& TIIF-S$\uparrow$
& TIIF-L$\uparrow$ \\
\midrule
Base
& 0.86 & 85.07 & 74.91 & 75.62 \\
AlphaGRPO
& 0.86 & 85.10 & 77.70 & 78.10 \\
\name{} (Ours)
& \textbf{0.89}
& \textbf{86.60}
& \textbf{80.68}
& \textbf{81.43} \\
\midrule

\multicolumn{5}{c}{\textbf{Image Editing: SenseNova-U1.5}} \\
\midrule
Reward Model
& ImgEdit
& GEdit-EN
& GEdit-CN
& RM Size \\
\midrule
Base
& 4.30 & 8.20 & 8.19 & -- \\
EditScore
& 4.51 & 8.31 & \textbf{8.37} & 72B (8$\times$) \\
\name{} (Ours)
& \textbf{4.52}
& \textbf{8.33}
& 8.34
& \textbf{9B (1$\times$)} \\
\bottomrule
\end{tabular}%
}
\end{wraptable}

\paragraph{Image Editing.}
We further compare \name with EditScore-72B~\citep{luo2026editscore} on SenseNova-U1.5 under identical training settings. 
Despite using only 9B parameters, one-eighth of EditScore-72B, \name achieves comparable overall performance and slightly higher results on ImgEdit (4.52 vs.\ 4.51) and GEdit-Bench-EN (8.33 vs.\ 8.31). 
Figure~\ref{fig:NEO_Compare} further examines the evolution of within-group reward variation during optimization. 
At training step 300, the reward standard deviation decreases by only 14.9\% with \name, compared with 50.5\% with EditScore-72B, indicating that \name retains greater reward variation among sampled candidates as training progresses. 
Such variation provides more differentiated signals for group-relative advantage estimation. 
Together, these results show that \name can provide effective reward guidance with substantially fewer parameters, enabling further optimization of an already strong image-editing model.

%% file: Sec/conclusion.tex
\section{Conclusion}

We introduce \textbf{Thinking Reward Model (\name)} under the \textbf{Think Before You Score} paradigm for visual reward modeling. 
\name constructs case-adaptive rubrics before scoring, enabling a shared evaluation process with flexible case-level criteria. 
We train \name through structured SFT followed by \textbf{\PDGRPO{}}, which leverages pairwise preferences to improve fine-grained discrimination while mitigating score polarization. 
Experiments demonstrate strong reward-modeling performance across image generation and editing, while \name{}-guided reinforcement learning consistently improves diverse generative models across architectures and scales.

%% file: Sec/appendix.tex
\section{Appendix}
\subsection{Data Details}
\label{data}
\subsubsection{Supervised Fine-Tuning Data}
\label{app:sft_data}
We construct approximately 48K cases for supervised fine-tuning, including around 20K image-generation cases and 28K image-editing cases. The data are designed to expose the model to diverse evaluation requirements and failure modes across different visual generation tasks. All collected cases are subsequently processed by the two-stage annotation pipeline described in the main paper to obtain structured supervision for training \name{}.

\paragraph{Image Generation Data.}
For image generation, we collect cases from diverse benchmarks and datasets, including Qwen-Image-Bench~\citep{li2026qwen}, EvalMuse~\citep{han2024evalmuse}, T2I-CoReBench~\citep{li2026easier}, AGIQA-3K~\citep{li2023agiqa}, HandEval~\citep{wang2025handeval}, PosterCraft/Poster100K~\citep{chen2025postercraft}, etc. These sources provide complementary coverage of prompt alignment, text rendering, reasoning, perceptual quality, and aesthetics. We reorganize their original annotations into the compact taxonomy summarized in Table~\ref{tab:sft_generation_categories}, merging closely related categories to obtain balanced coverage rather than treating every original benchmark label as an independent task.

\begin{figure*}[t]
    \centering
    \begin{subfigure}[t]{0.48\textwidth}
        \centering
        \includegraphics[width=\linewidth]{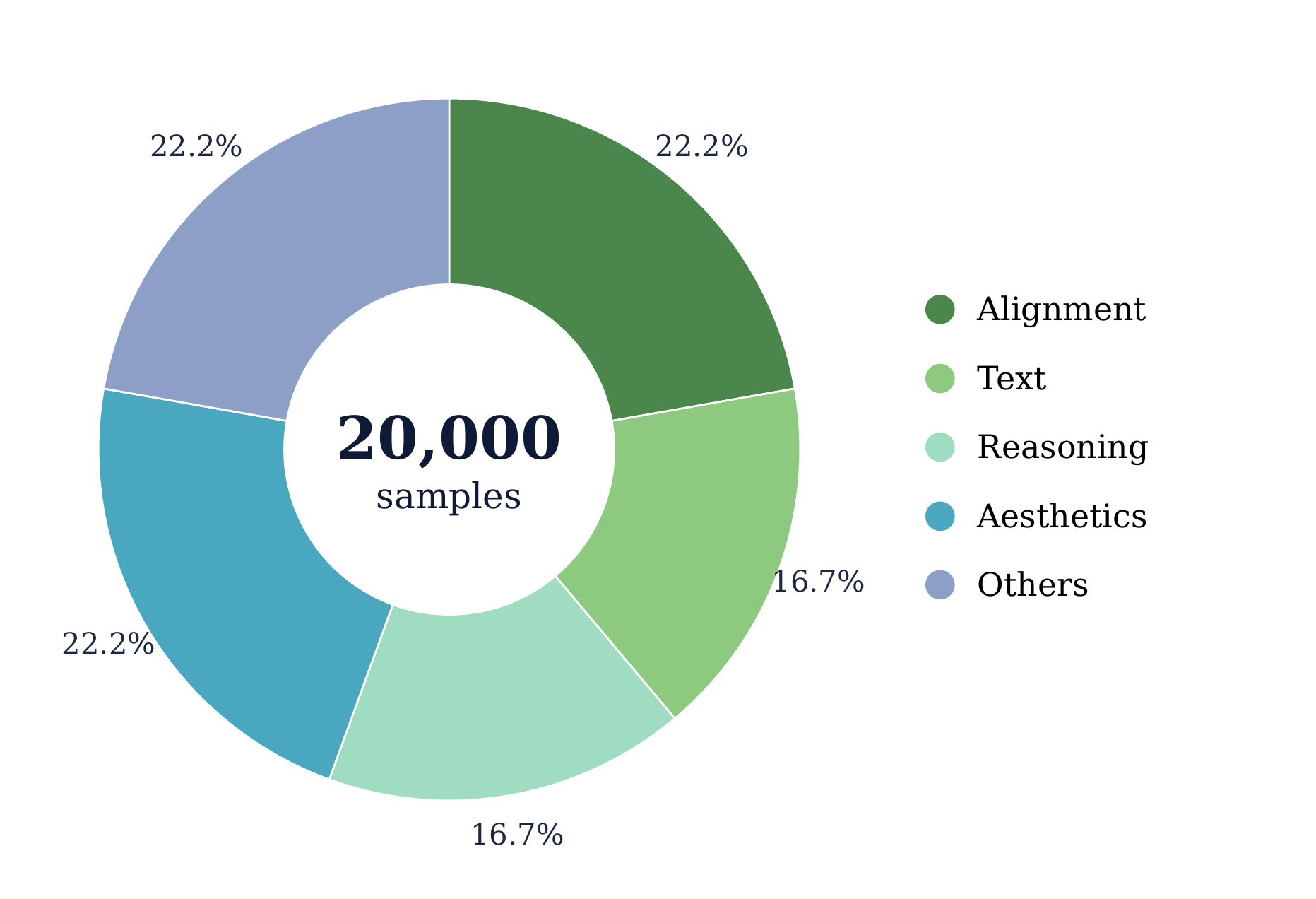}
        \caption{Image-generation SFT data distribution.}
        \label{fig:generation_sft_distribution}
    \end{subfigure}
    \hfill
    \begin{subfigure}[t]{0.48\textwidth}
        \centering
        \includegraphics[width=\linewidth]{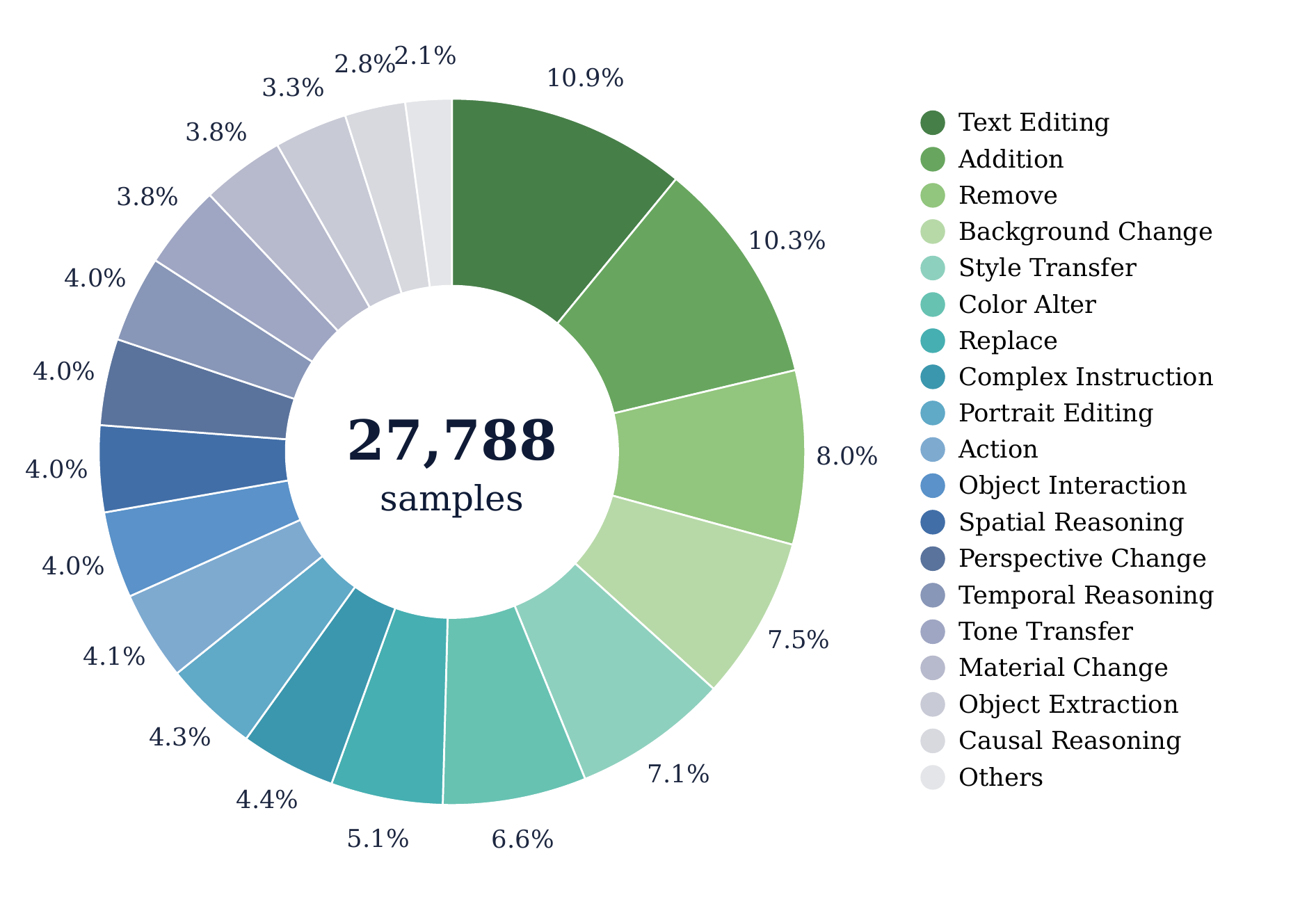}
        \caption{Image-editing SFT data distribution.}
        \label{fig:editing_sft_distribution}
    \end{subfigure}
    \caption{
    \textbf{Task distributions of the SFT data.}
    Left: the image-generation SFT data are organized into fine-grained task categories under the compact taxonomy in Table~\ref{tab:sft_generation_categories}.
    Right: the image-editing SFT data span 19 fine-grained task categories under the taxonomy in Table~\ref{tab:sft_editing_categories}.
    Together, these distributions illustrate the balanced and diverse coverage of our SFT data across visual generation tasks.
    }
    \label{fig:sft_data_distribution}
\end{figure*}

Beyond general alignment and visual quality, we introduce several focused subsets to complement generic text-to-image evaluation. Human-centric cases emphasize portrait realism and fine-grained anatomical defects, including facial structure, hands, and body anomalies, where localized errors can strongly affect perceptual quality despite largely correct global semantics. We further include poster-oriented examples involving layout, typography, and text--image interaction, as well as Chinese and multilingual text-rendering cases. Together with dedicated short-text, long-text, and compositional text cases, these subsets broaden the coverage of legibility, placement, and visual--text consistency. Reasoning-oriented data additionally cover logical, causal, behavioral, and compositional requirements. Overall, the generation data span prompt fidelity, perceptual quality, structural coherence, and fine-grained visual defects. These categories are used for data construction and balancing only and do not prescribe fixed rubrics during annotation or inference. The resulting task distribution is shown in Figure~\ref{fig:sft_data_distribution}.

\begin{table}[htbp]
\centering
\caption{\textbf{Taxonomy of the image generation SFT data.} We organize diverse generation cases into five complementary groups for data balancing; these categories do not serve as fixed evaluation rubrics.}
\label{tab:sft_generation_categories}
\small
\setlength{\tabcolsep}{6pt}
\renewcommand{\arraystretch}{1.08}
\begin{tabular}{p{0.17\linewidth}p{0.34\linewidth}p{0.37\linewidth}}
\toprule
\textbf{Group} & \textbf{Categories} & \textbf{Evaluation Focus} \\
\midrule
Alignment & subject, attribute, count space, action scene & Subject presence, attribute binding, counting, spatial relations, actions, and scene conditions \\
Text & few, many, combo & Short text, long text, and text rendering combined with other visual requirements \\
Reasoning & logic, causal, analog & Logical, causal, behavioral, analogical, generalization, and procedural reasoning \\
Aesthetics & quality, color, view, detail & Overall quality, color, composition/viewpoint, and fine-grained detail \\
Others & portrait, body anomaly, poster, multiling text & Portrait realism, body/hand anomalies, poster composition, and multilingual text rendering \\
\bottomrule
\end{tabular}
\end{table}

\paragraph{Image Editing Data.}
\label{editing}
For image editing, we construct the SFT data from two
complementary sources. First, we collect cases from EditScore
and filter the original source-image--instruction pairs using
Qwen3.5-397B-A17B~\citep{qwen3.5}.
The filtering process checks whether the referenced objects
are present in the source image and whether the requested
operations are applicable and clearly specified.
Pairs with incompatible or ambiguous instructions are excluded.
We additionally rebalance the retained cases according to
their original EditScore ratings, reducing the sampling
proportions of the highest- and lowest-scoring examples
while increasing the proportion of examples with intermediate
scores. This strategy places greater emphasis on intermediate
quality levels while retaining coverage of both high- and
low-quality outputs

Second, we jointly curate editing conditions and candidate
outputs to provide informative supervision for reward-model
training. We select source images and editing instructions
from EditCompass~\citep{bai2026edit},
UniREdit-Bench~\citep{han2026unireditbench},
UNIC-Bench~\citep{ye2026unicedit},
RISE-Bench~\citep{zhao2026envisioning},
KRIS-Bench~\citep{wu2026kris}, and related editing benchmarks
to broaden task coverage. We emphasize challenging edits
involving spatial relations, object interactions, perspective
changes, temporal or causal reasoning, and multiple constraints.
For each task category, we use rankings on relevant benchmarks
to select editing models spanning different capability levels
and sample their outputs. Taking spatial editing as an example, we select object movement, object swapping, and relation change tasks from multiple benchmarks and collect candidate outputs under the same source images and editing instructions.
Specifically, for Object Movement and Object Swap tasks from
Edit-Compass, we sample outputs from models including BAGEL~\cite{zhao2026envisioning},
FLUX.2-dev~\cite{flux-2-2025}, and Gemini 3.1 Flash Image Preview.
For Relation Change tasks from GEdit-Bench-v2, we use models
including FLUX.1 Kontext-Dev~\cite{flux2024}, Qwen-Image-Edit-2509~\cite{wu2025qwen}, and
Nano-Banana-Pro.
This construction combines diverse editing tasks with variation
in candidate quality, capturing differences in instruction
following and visual consistency across models under matched
task conditions.
As summarized in Table~\ref{tab:sft_editing_categories}, the resulting data cover both common appearance-level modifications and reasoning-intensive editing tasks. Compared with image generation, image editing additionally requires evaluating consistency with the source image. The reward model must therefore jointly assess instruction fulfillment, visual quality, and preservation of content outside the intended edit region. As in the generation setting, these task categories are used to diversify the training distribution rather than to define fixed evaluation rubrics. The corresponding task distribution is also shown in
Figure~\ref{fig:sft_data_distribution}.

\begin{table}[t]
\centering
\caption{\textbf{Taxonomy of the image editing SFT data.} The collected cases cover both direct visual modifications and reasoning-intensive editing operations.}
\label{tab:sft_editing_categories}
\small
\setlength{\tabcolsep}{7pt}
\renewcommand{\arraystretch}{1.08}
\begin{tabular}{p{0.32\linewidth}p{0.58\linewidth}}
\toprule
\textbf{Category} & \textbf{Evaluation Focus} \\
\midrule
Addition & Accurate insertion and natural integration \\
Remove & Complete removal and seamless region restoration \\
Replace & Accurate replacement and natural integration \\
Text Editing & Text accuracy, legibility, and visual consistency \\
Background Change & Background accuracy and foreground preservation \\
Style Transfer & Target style alignment and content preservation \\
Color Alter & Targeted recoloring and texture preservation \\
Portrait Editing & Requested appearance changes and identity preservation \\
Complex Instruction & Multi-constraint instruction following \\
Material Change & Material and surface-property modification \\
Tone Transfer & Global or regional tone transformation \\
Action & Modification of subject actions \\
Object Interaction & Interaction and relational consistency between objects \\
Spatial Reasoning & Spatial relationships and geometric consistency \\
Causal Reasoning & Causality-aware modification of scene content \\
Object Extraction & Extraction and preservation of the target object \\
Perspective Change & Viewpoint and perspective transformation \\
Temporal Reasoning & Temporally conditioned scene modification \\
Size Adjustment & Relative or absolute object-scale modification \\
Emotion Change & Facial expression and emotional-state modification \\
\bottomrule
\end{tabular}
\end{table}

\paragraph{Data Balancing.}
We balance the SFT data across image generation and editing, task categories, candidate sources, and quality levels. As shown in Figure~\ref{fig:sft_data_distribution}, the resulting
data exhibit broad and structured coverage for both tasks. For image generation, the 20K cases cover fine-grained categories spanning prompt alignment, text rendering, reasoning, aesthetics,
and several focused subsets such as portrait realism, poster composition, and multilingual text.
For image editing, the 27,788 cases span 19 fine-grained task categories, covering both common appearance-level modifications and reasoning-intensive editing operations.
We additionally collect candidate outputs from models with different capability levels to cover a broad quality range. Overall, the approximately 48K SFT cases encompass diverse task
conditions and candidate qualities, while the evaluation criteria for each individual case are generated adaptively through the annotation pipeline described in the main paper.

\subsubsection{Reinforcement Learning Data}
\label{app:rl_data}
We further construct pairwise preference data for reinforcement learning to improve the fine-grained discrimination ability of \name{}(SFT). The RL data cover diverse task content and comparison difficulty, including both clearly distinguishable candidates and challenging pairs with relatively small quality differences.

\paragraph{Image Generation Data.}
For image generation, we collect preference data from Open Image Preferences (OIP), EvalMuse~\citep{han2024evalmuse}, and HPDv3++~\citep{liu2026hpsv3++}. We select candidate pairs with diverse prompt content and visual characteristics to ensure broad coverage of generation cases. We additionally balance the data across different difficulty levels. Following Sec.~4.1, the score difference between two candidates is used as a proxy for comparison difficulty, with larger gaps corresponding to easier pairs and smaller gaps corresponding to harder pairs. We retain easy, medium, and hard examples so that the RL data contain both clear preference signals and fine-grained comparisons between candidates of similar quality.

\paragraph{Image Editing Data.}
For image editing, we construct preference pairs for reward-model
RL from the high-quality data curated for cold-start SFT.
We select diverse editing tasks and pair candidate outputs
conditioned on the same source image and instruction,
enabling quality comparisons under matched task conditions.
The resulting pairs cover diverse editing operations and
visual content, with quality differences in instruction
fulfillment, visual fidelity, spatial correctness, and
preservation of content unrelated to the requested edit.
Following the score-gap criterion used for image generation,
we adjust the sampling proportions across easy, medium,
and hard comparisons. By including both clear preferences
and comparisons between candidates of similar quality,
we aim to encourage the model to identify subtle editing
errors and perform more detailed quality analysis,
leading to more accurate and reliable reward predictions.

\subsection{Training Details}
\label{supp:training_details}
\subsubsection{Supervised Fine-Tuning of the Reward Model}
\label{app:sft_training}

Although the structured annotations are constructed through the two-stage annotation pipeline described in Sec.~\ref{sft}, we do not train the reward model in separate stages. Instead, we perform a single-stage supervised fine-tuning procedure on the complete structured responses, such that the model jointly learns rubric generation, rubric-level evaluation, dimension-level reasoning, and final score prediction.
We initialize \name{}(SFT) from Qwen3.5-9B and fine-tune it using ms-swift with LoRA. The LoRA rank and scaling factor are set to 32 and 64, respectively, with a dropout rate of 0.05. LoRA adapters are applied to all linear layers, while the visual tower is frozen throughout training. We use BF16 precision together with DeepSpeed ZeRO-2. The maximum sequence length is set to 8,192 tokens, and the number of image tokens is capped at 1,024. Training samples are grouped by sequence length to improve computational efficiency.

The SFT data contain approximately 20K image generation cases and 28K image editing cases. Each training response follows the three-dimensional checklist protocol described in Sec.~\ref{rubric}, containing case-adaptive rubrics and rubric-level judgments under the three high-level evaluation dimensions, followed by dimension-level reasons and a final pointwise score. The final scores are re-annotated after calibration with Gemini. For image generation, we find that scores from a single teacher model provide relatively limited separation among candidates of similar quality. We therefore additionally incorporate score calibration from other evaluator models to improve the discriminability and robustness of the final-score supervision. The training and validation sets are split at the prompt level, ensuring that samples associated with the same prompt do not appear in both splits.

We train the model on 8 GPUs with an effective batch size of 64, using a per-device batch size of 2 and gradient accumulation over 4 steps. The learning rate is set to $1\times10^{-5}$ with a cosine learning-rate schedule and a warmup ratio of 0.1. Training lasts for 2 epochs. We select the intermediate checkpoint with the best validation performance and use it as the initial policy for subsequent reinforcement learning.

\subsubsection{Reinforcement Learning of the Reward Model}

Starting from \name{}(SFT), we perform pairwise preference optimization using \PDGRPO{}. Each preference pair consists of two candidate images under the same task condition. The two candidates are processed independently, and we sample 8 evaluation responses for each candidate, resulting in 16 rollouts for each preference pair. This dual-group construction provides relative supervision between the preferred and worse candidates while preserving the pointwise evaluation interface of the reward model.
The RL training set contains approximately 4K preference pairs, corresponding to 8K candidate samples. We additionally construct a validation set of 300 challenging pairs that are incorrectly ranked by the SFT model, allowing validation to focus on preference cases for which further optimization is most useful. All final scores used for reward construction are normalized before computing the pairwise reward. Following the formulation in Sec.~\ref{optimization}, we use a hard-margin pairwise reward. For each rollout, the preference reward is determined by whether its normalized final score is sufficiently separated from the mean score of the opposite rollout group:
\begin{equation}
r_{\mathrm{pref}}(y)=
\begin{cases}
\mathbb{I}[s(y)-\mu^{-}>m], & y\in G^{+},\\
\mathbb{I}[\mu^{+}-s(y)>m], & y\in G^{-}.
\end{cases}
\end{equation}
Here, $\mu^{+}$ and $\mu^{-}$ denote the mean normalized scores of the preferred and worse rollout groups, respectively, and $m$ is the separation margin. For image editing, we set $m=0$, while for image generation, where preference pairs contain more ties and near-ties, we set $m=0.05$ to require a clearer score separation. Since all final scores are normalized before reward computation, both margins are defined on the normalized score scale. We additionally use a format reward of $0.1$ to encourage valid structured outputs.

We optimize the policy using LoRA with rank 32 and scaling factor 64. LoRA adapters are applied to all linear layers except those in the visual tower and linear-attention modules, and training is performed in BF16 precision. We use task-specific optimization hyperparameters: for image generation, the learning rate is set to $2\times10^{-6}$ with a KL regularization coefficient of $0.04$; for image editing, we use a higher learning rate of $5\times10^{-6}$ and a smaller KL coefficient of $0.003$. Both settings are optimized for 3 epochs, with the checkpoint achieving the best validation performance selected for the final evaluation.
The global training batch contains 32 candidate samples, corresponding to 16 preference pairs, with a mini-batch size of 8. During rollout generation, the sampling temperature is set to 0.7, while validation uses deterministic decoding with temperature 0. The maximum prompt and response lengths are 32,768 and 4,096 tokens, respectively. All reinforcement learning experiments are conducted on 8 GPUs, with vLLM used for rollout generation.

\subsubsection{Reinforcement Learning for Image Generation}
We optimize BAGEL, FLUX.1-dev, and SD3.5-M using FlowGRPO,
with \name{} providing the reward signal.
All three models are trained using LoRA.
The model-specific training configurations are detailed below.

\noindent\hspace*{2em}\textbf{BAGEL.}
We optimize BAGEL-7B-MoT for text-to-image generation using
FlowGRPO with \name{} as the sole reward model.
Training uses 8 GPUs, LoRA with rank 64 and scaling factor
128, BF16 mixed precision, group size $G=24$, and a
per-device batch size of 3.
We use AdamW with a learning rate of $10^{-4}$, weight
decay of $10^{-4}$, and gradient norm clipping at 1.0.
Advantages are normalized using the global standard deviation
and clipped at 5.0. The policy clipping range is $10^{-4}$,
with $\beta=0$. EMA decay is set to 0.9.
Rollouts use 10 sampling steps at a resolution of
$512\times512$, with CPS dynamics, noise level 0.8,
scheduler shift 3.0, and three stochastic steps sampled
from the first six steps. The text guidance scale is
set to 1.0.

\noindent\hspace*{2em}\textbf{FLUX.1-dev.}
We optimize FLUX.1-dev for text-to-image generation using
FlowGRPO with \name{} as the sole reward model.
Training uses LoRA with rank 64 and scaling factor 128,
BF16 mixed precision, group size $G=24$, and a per-device
batch size of 3.
We use AdamW with a learning rate of $3\times10^{-4}$,
weight decay of $10^{-4}$, and gradient norm clipping at 1.0.
Global standard-deviation normalization is disabled, and
advantages are clipped at 5.0. The policy clipping range
is $10^{-4}$, with $\beta=0$. EMA decay is set to 0.9.
Rollouts use 10 sampling steps at a resolution of
$512\times512$, with CPS dynamics, noise level 0.7,
guidance scale 3.5, and two stochastic steps sampled
from the first four steps.

\noindent\hspace*{2em}\textbf{SD3.5-M.}
We optimize Stable Diffusion 3.5 Medium for text-to-image
generation using an AlphaGRPO-style FlowGRPO configuration
with \name{} as the sole reward model.
Training uses LoRA with rank 32 and scaling factor 64,
BF16 mixed precision, group size $G=24$, and a per-device
batch size of 6.
We use AdamW with a learning rate of $3\times10^{-4}$,
weight decay of $10^{-4}$, and gradient norm clipping at 1.0.
Advantages are normalized using the global standard deviation
and clipped at 5.0. The policy clipping range is $10^{-4}$,
with $\beta=0$. EMA decay is set to 0.99.
Rollouts use 10 sampling steps at a resolution of
$512\times512$, with CPS dynamics, noise level 0.8,
guidance scale 4.5, and two stochastic steps sampled
from step indices $\{1,2,3,4,5\}$.

\subsubsection{Reinforcement Learning for Image Editing}
We optimize image-editing models using FlowGRPO,
with \name{} providing the reward signal and LoRA
used for parameter-efficient fine-tuning.
The model-specific training configurations are detailed below.

\noindent\hspace*{2em}\textbf{BAGEL.}
We optimize BAGEL-7B-MoT using FlowGRPO with \name{} as
the sole reward model. Training uses LoRA with rank 64
and scaling factor 128, BF16 mixed precision, group size
$G=24$, and a per-device batch size of 3.
We use AdamW with a learning rate of $10^{-4}$, weight
decay of $10^{-4}$, and gradient norm clipping at 1.0.
Advantages are normalized using the global standard deviation
and clipped at 5.0. The policy clipping range is $10^{-4}$,
with $\beta=0$. EMA decay is set to 0.9.
Rollouts use 10 sampling steps with CPS dynamics, noise
level 0.8, scheduler shift 3.0, and three stochastic steps
sampled from the first six steps.
Output dimensions follow the source image, with total
pixel counts in $[512^2,1024^2]$.
Text and image guidance scales are both set to 1.0.

\noindent\hspace*{2em}\textbf{Flux-Kontext.}
We optimize FLUX.1 Kontext-dev using FlowGRPO with \name{}
as the sole reward model. Training uses 8 GPUs, LoRA with
rank 64 and scaling factor 128, BF16 mixed precision,
group size $G=16$, and a per-device batch size of 2.
We use AdamW with a learning rate of $1.5\times10^{-4}$,
weight decay of $10^{-4}$, and gradient norm clipping at 1.0.
Advantages are normalized using the global standard deviation
and clipped at 5.0. The policy clipping range is $10^{-4}$,
with $\beta=0$. EMA decay is set to 0.9.
Rollouts use 10 sampling steps with CPS dynamics, noise
level 0.9, guidance scale 2.5, and two stochastic steps
sampled from the first four steps.
Both image and conditioning resolutions are set to 512.

\noindent\hspace*{2em}\textbf{FLUX.2-Klein-4B.}
We optimize FLUX.2 Klein Base 4B using FlowGRPO with \name{}
as the sole reward model. Training uses LoRA with rank 64
and scaling factor 128, BF16 mixed precision, group size
$G=16$, and a configured per-device batch size of 2.
We use AdamW with a learning rate of $3\times10^{-4}$,
weight decay of $10^{-4}$, and gradient norm clipping at 1.0.
Advantages are normalized using the global standard deviation
and clipped at 5.0. The policy clipping range is $10^{-5}$,
with $\beta=0$. EMA decay is set to 0.9.
Rollouts use 20 sampling steps with CPS dynamics, noise
level 0.7, guidance scale 4.0, and three stochastic steps
sampled from the first six steps.
Output dimensions follow the source image, with the total
pixel count capped at $384^2$.

\noindent\hspace*{2em}\textbf{FLUX.2-Klein-9B.}
We optimize FLUX.2 Klein Base 9B using FlowGRPO with \name{}
as the sole reward model. Training uses LoRA with rank 64
and scaling factor 128, BF16 mixed precision, group size
$G=16$, and a configured per-device batch size of 2.
We use AdamW with a learning rate of $3\times10^{-4}$,
weight decay of $10^{-4}$, and gradient norm clipping at 1.0.
Global standard-deviation normalization is disabled, and
advantages are clipped at 5.0. The policy clipping range
is $10^{-4}$, with $\beta=0$. EMA decay is set to 0.9.
Rollouts use 20 sampling steps with CPS dynamics, noise
level 0.7, guidance scale 4.0, and three stochastic steps
sampled from the first six steps.
Output dimensions follow the source image, with the total
pixel count capped at $384^2$.

\subsection{Additional Analyses}
\label{sec:additional_analyses}
\subsubsection{Relative Advantages of Pointwise and Pairwise Reward Modeling}
\label{sec:pointwise_pairwise}
\paragraph{Definitions.}
Let $c$ denote the task condition, consisting of a text prompt
for image generation or a source image and an editing instruction
for image editing. Pairwise reward modeling jointly evaluates
two candidate outputs, $x_A$ and $x_B$, under the same condition
$c$ and predicts their relative preference, indicating which
candidate better satisfies the evaluation requirements.
Pointwise reward modeling instead evaluates each candidate
$x$ independently and assigns a scalar score $r(c,x)$ that
reflects its quality under the given condition. A pairwise
preference can then be inferred by comparing the scores of
two candidates.

\paragraph{Comparative Analysis.}
We compare the two formulations in terms of preference
consistency and their use in policy optimization.

\textit{Preference consistency.}
A reliable pairwise judgment should preserve the preferred
candidate when the presentation order is reversed.
However, Table~\ref{tab:pairwise_order_consistency} reveals
substantial inconsistency under order reversal, particularly
among the smaller models evaluated. For example, Qwen3.5-9B
exhibits an inconsistency rate of 54.7\%, despite similar
forward and reverse accuracies. When binary preference
accuracy is averaged across both orders, each inconsistent
pair contributes exactly 50\% accuracy. Aggregate accuracy
alone therefore does not fully characterize the reliability
of preference judgments. If used directly for policy
optimization, such judgments could introduce contradictory
reward feedback for the same candidate pair.

\begin{table}[t]
\centering
\caption{\textbf{Pairwise preference consistency under order reversal
on MMRB2.}
Forward and reverse evaluations present the same candidates
in opposite orders. Consistency is measured after mapping
predictions back to candidate identities.
All values are percentages.}
\label{tab:pairwise_order_consistency}
\setlength{\tabcolsep}{5pt}
\renewcommand{\arraystretch}{1.12}
\begin{tabular}{lccccc}
\toprule
\textbf{Model} &
\textbf{Avg. Acc.} &
\textbf{Forward Acc.} &
\textbf{Reverse Acc.} &
\textbf{Consistent} &
\textbf{Inconsistent} \\
\midrule
Qwen3-VL-8B    & 62.0 & 63.9 & 59.9 & 55.9 & 44.1 \\
Qwen3.5-9B     & 51.4 & 51.3 & 51.5 & 45.3 & 54.7 \\
Qwen2.5-VL-72B & 65.8 & 65.8 & 65.8 & 74.6 & 25.5 \\
\bottomrule
\end{tabular}%
\vspace{-0.5cm}
\end{table}

\textit{Policy optimization.}
In group-based reinforcement learning, group size determines
the number of candidates sampled under each task condition.
Larger groups offer more opportunities to explore diverse
outputs, but also increase the demand for reward computation.
For a group of $G$ candidates, pointwise scoring requires
$G$ independent evaluations, yielding a cost that scales
linearly with group size. In contrast, exhaustive pairwise
comparison requires $G(G-1)/2$ evaluations and thus incurs
quadratic cost, with further overhead if both presentation
orders are evaluated for consistency. Reference-based or
sparse comparisons reduce this cost, but make the resulting
rewards depend on the selected comparison partners.
Pointwise scoring therefore supports larger rollout groups
while controlling reward computation costs.
These considerations motivate our use of pointwise rewards
for policy optimization, together with pairwise preference
supervision to improve relative quality discrimination
during reward-model training.

\subsubsection{Analysis of Efficiency of \name{}}
Using \name{} as an MLLM-based reward model introduces
additional inference costs during policy optimization.
A synchronous rollout-then-reward workflow serializes image
generation and reward computation, while repeated processing
of the shared system prompt incurs redundant computation.
We deploy \name{} as a dedicated reward inference server
using vLLM and reduce these overheads through asynchronous
reward computation and shared-prefix caching.

\paragraph{Asynchronous Reward Computation.}
During rollout generation, completed samples are accumulated
and submitted asynchronously to the vLLM reward server once
a predefined batch threshold is reached.
Reward inference proceeds concurrently with subsequent
rollouts, and any remaining samples are submitted when
generation completes. All rewards are collected before
advantage computation and policy optimization.

Consider $K$ batches, with per-batch generation and reward
computation times denoted by $t_g$ and $t_r$, respectively.
Under an idealized two-stage pipeline with separate resources,
constant processing times, and negligible communication
overhead, the collection times are
\begin{align}
T_{\mathrm{sync}} &= K(t_g+t_r), \\
T_{\mathrm{async}} &= t_g+t_r+(K-1)\max(t_g,t_r).
\end{align}
Overlapping the two stages therefore saves
$(K-1)\min(t_g,t_r)$, with the largest relative benefit
when generation and reward computation have comparable costs.
This analysis excludes the subsequent policy update.

\paragraph{Shared-Prefix Caching.}
We enable prefix caching in the vLLM server to reuse the
cached states of the shared system prompt, which specifies
the evaluation procedure and output format.
On a cache hit, the corresponding prefill computation is
skipped, reducing redundant processing across reward requests.
Candidate-specific inputs and assessment outputs are
still processed independently.

\begin{figure*}[t]
    \centering
    \includegraphics[width=\textwidth]{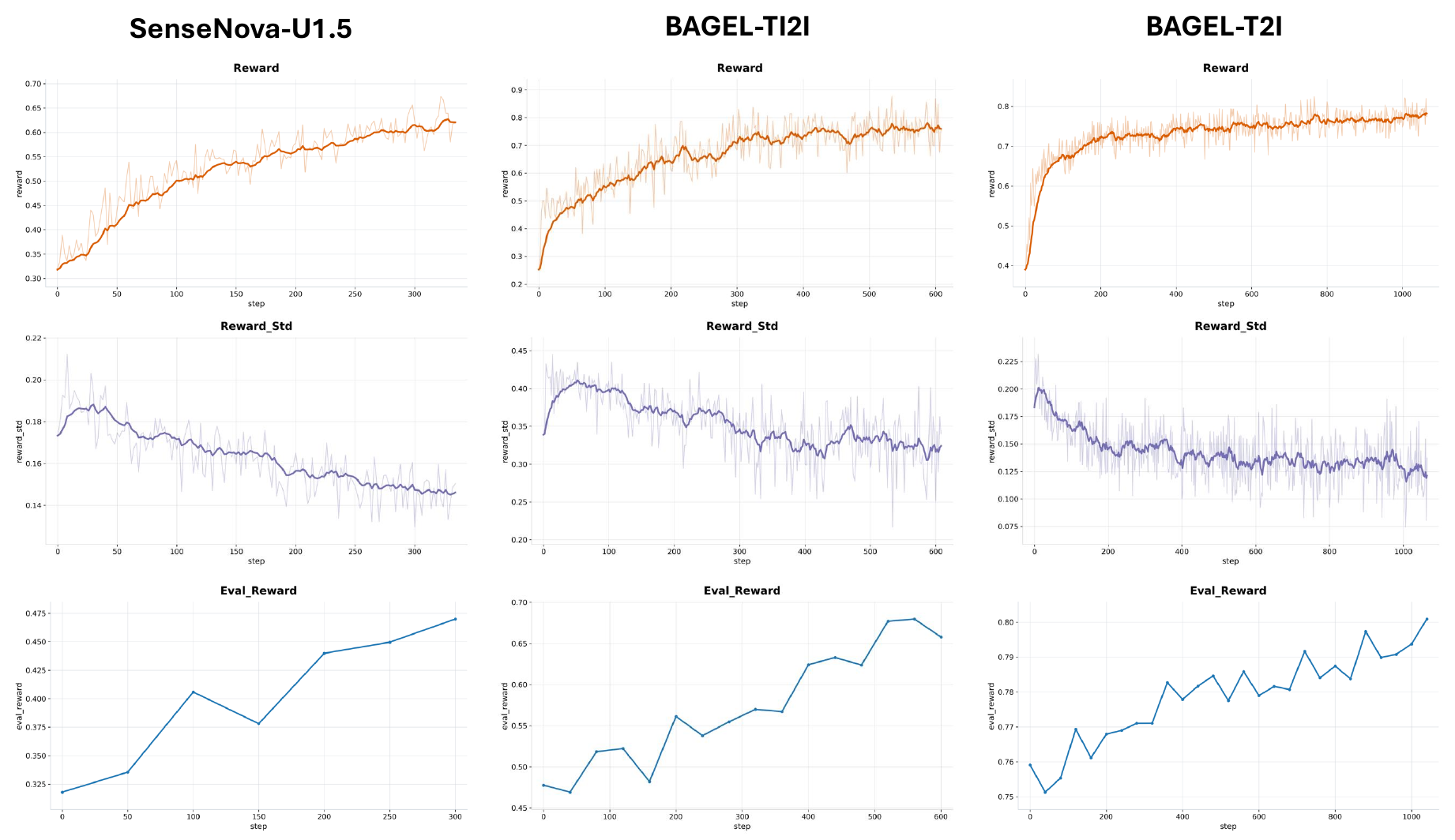}
    \caption{
    \textbf{Reward dynamics during \name{}-guided RL.}
    Training reward, reward standard deviation, and evaluation
    reward for SenseNova-U1.5 and BAGEL on image editing
    and BAGEL on image generation.
    }
    \label{fig:reward_curves}
\end{figure*}

\begin{figure*}[t]
    \centering
    \includegraphics[width=\textwidth]{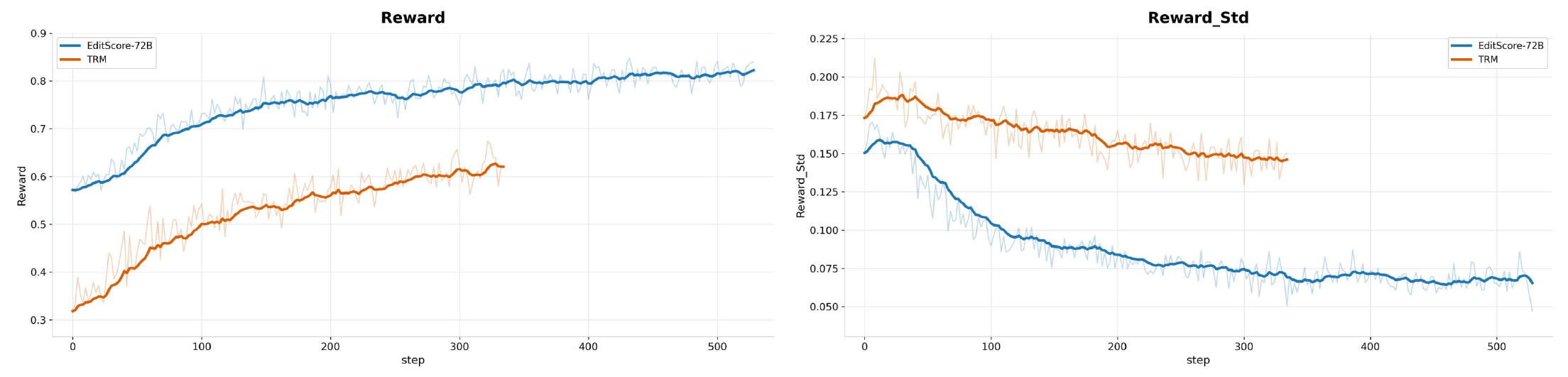}
    \caption{
    \textbf{Training dynamics with different reward models.}
    Comparison of mean reward and reward standard deviation
    during RL guided by \name{} and the larger
    EditScore-72B reward model.
    }
    \label{fig:NEO_Compare}
\end{figure*}

\begin{figure*}[t]
    \centering
    \includegraphics[width=0.75\textwidth]{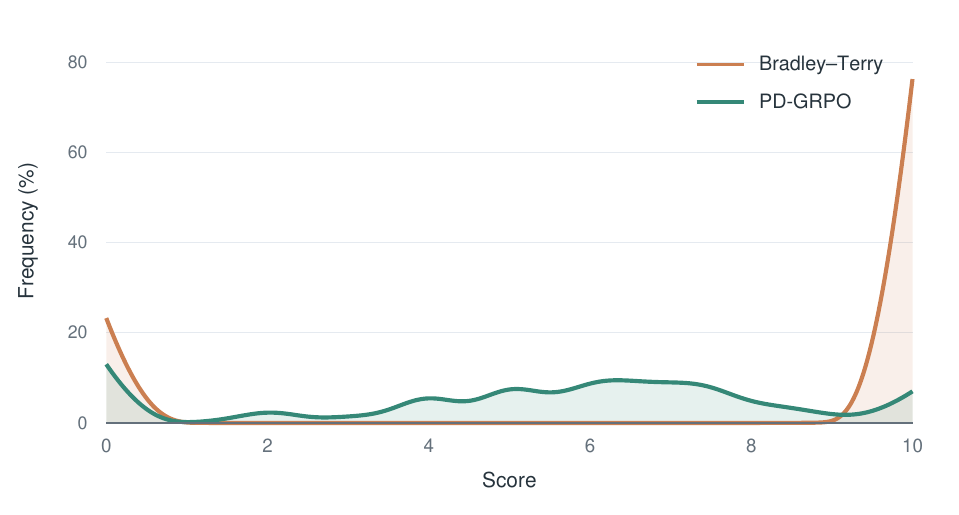}
    \caption{
    \textbf{Comparison of score distributions between PD-GRPO and Bradley--Terry optimization.}
    }
    \label{fig:compare_score}
\end{figure*}

\subsection{Additional Experimental Results}
\label{additional}
\subsubsection{Qualitative Results of \name{}-Guided Optimization}

To complement the quantitative results in Sec.~\ref{RL}, we provide qualitative comparisons between the original generation models and their counterparts optimized with \name{}. We include examples from both image editing and image generation, covering multiple model families. These examples illustrate how the quantitative improvements translate into visible changes in instruction following, content preservation, visual quality, and prompt alignment.

\begin{figure}[t]
    \centering
    \includegraphics[width=\textwidth]{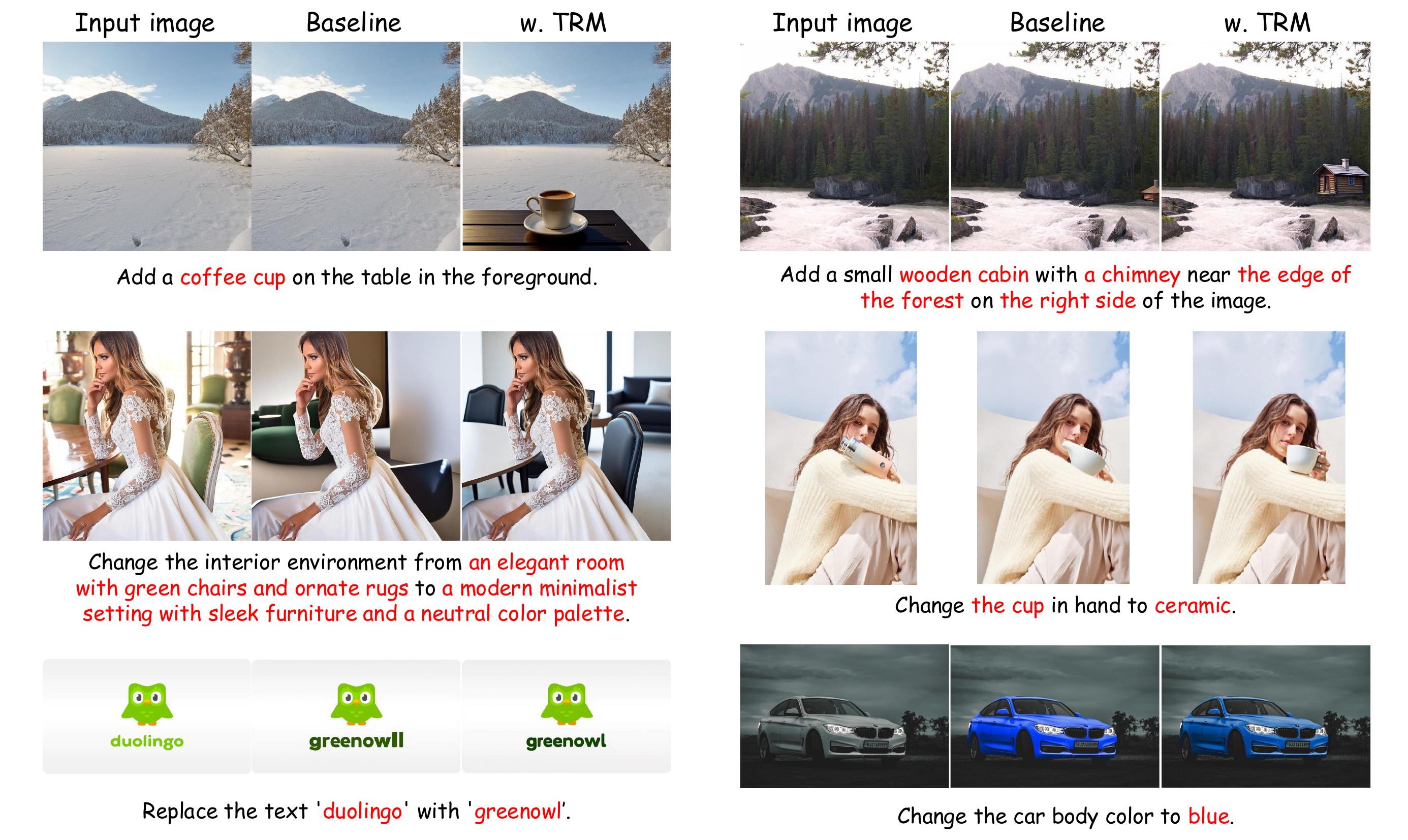}
    \caption{
    \textbf{Qualitative results of \name{}-guided optimization on BAGEL for image editing.} For each example, we show the input image, the output of the original model,and the output after reinforcement learning with \name{} as the reward. \name{}-guided optimization improves instruction following and edit quality across diverse editing tasks while preserving unrelated image content.
    }
    \label{fig:qual_edit_bagel}
\end{figure}

\begin{figure}[t]
    \centering
    \includegraphics[width=\textwidth]{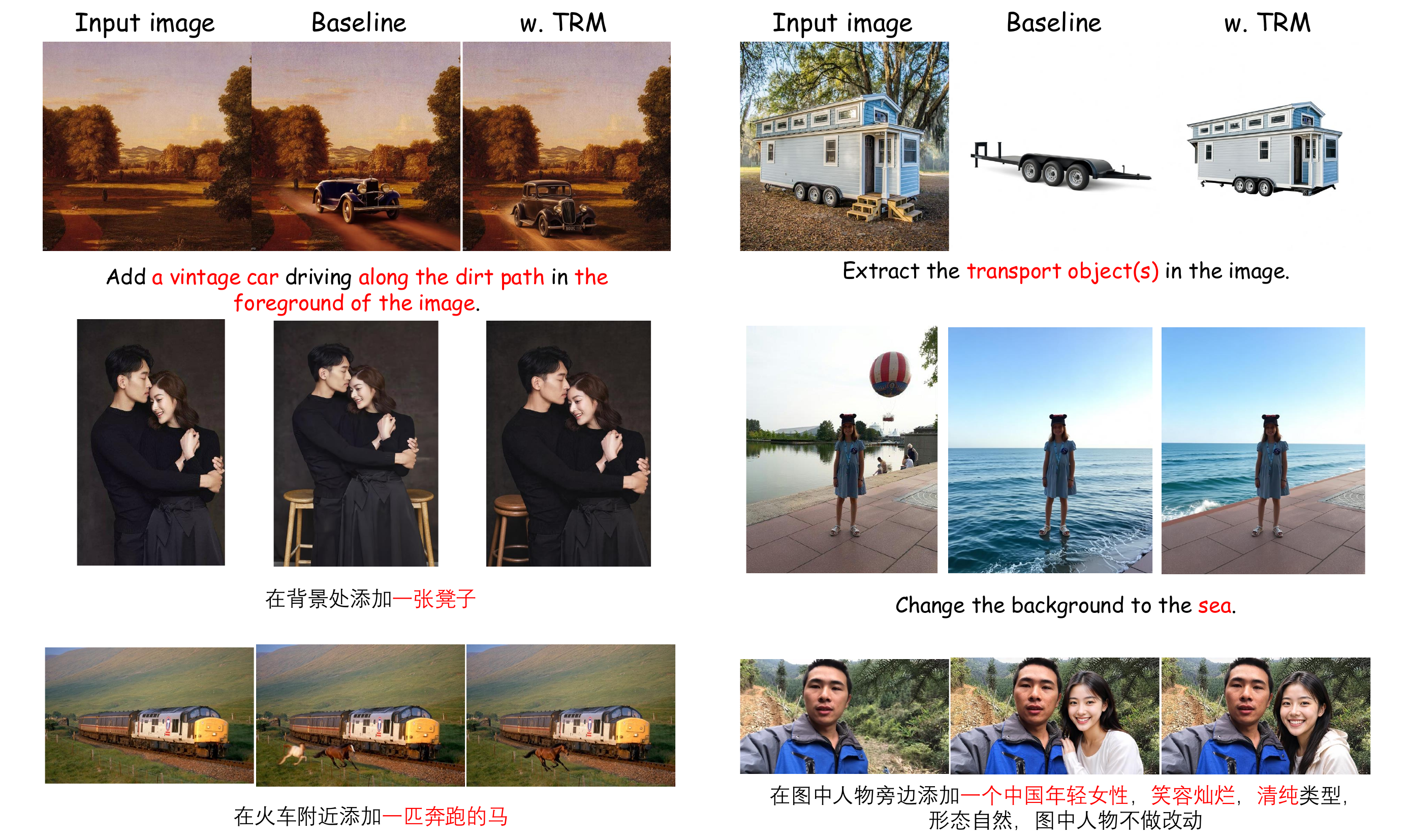}
    \caption{\textbf{Qualitative results of \name{}-guided optimization on SenseNova-U1.5 for image editing.} We compare outputs from the original model and its \name{}-optimized counterpart across diverse editing instructions. \name{}-guided optimization produces more accurate edits while maintaining visual consistency with the input image.}
    \label{fig:qual_edit_sensenova}
    \vspace{-0.5cm}
\end{figure}

\paragraph{Image Editing.}
Figures~\ref{fig:qual_edit_bagel} and~\ref{fig:qual_edit_sensenova}
show representative results for BAGEL and SenseNova-U1.5, respectively.
Compared with the corresponding base models, \name{}-guided optimization more
reliably applies the requested edits while preserving unrelated image content.
The improvements cover diverse operations, including object insertion,
attribute modification, text replacement, background changes, object
extraction, and compositional edits.

\begin{figure}[t]
    \centering
    \includegraphics[width=\textwidth]{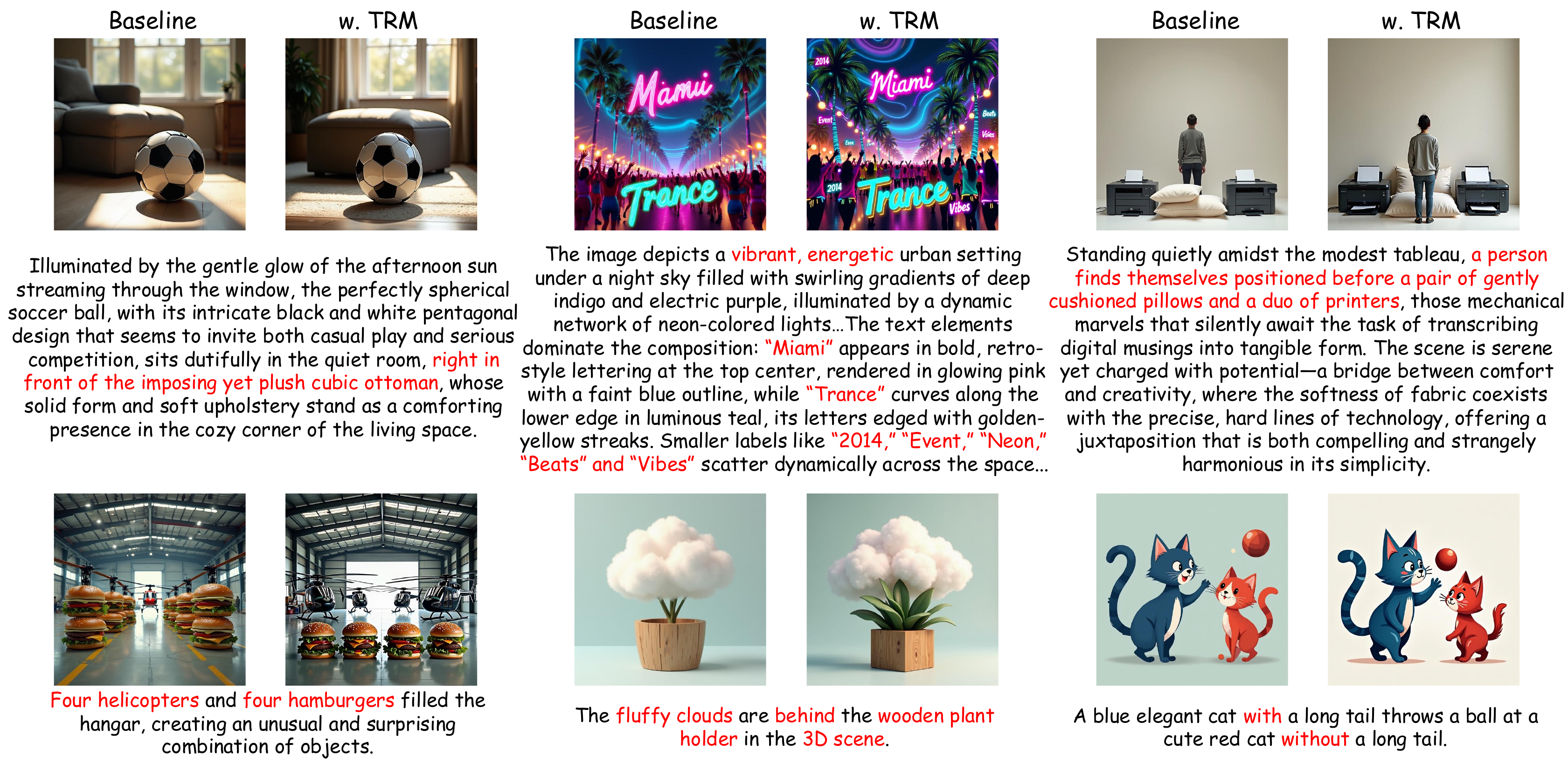}
    \caption{ \textbf{Qualitative results of \name{}-guided optimization on BAGEL for image generation.} Given the same text prompts, we compare generations from the original model and its \name{}-optimized counterpart. \name{}-guided optimization improves prompt alignment and fine-grained visual fidelity across diverse generation cases.
}
    \label{fig:qual_gen_bagel}
\end{figure}

\begin{figure}[t]
    \centering
    \includegraphics[width=\textwidth]{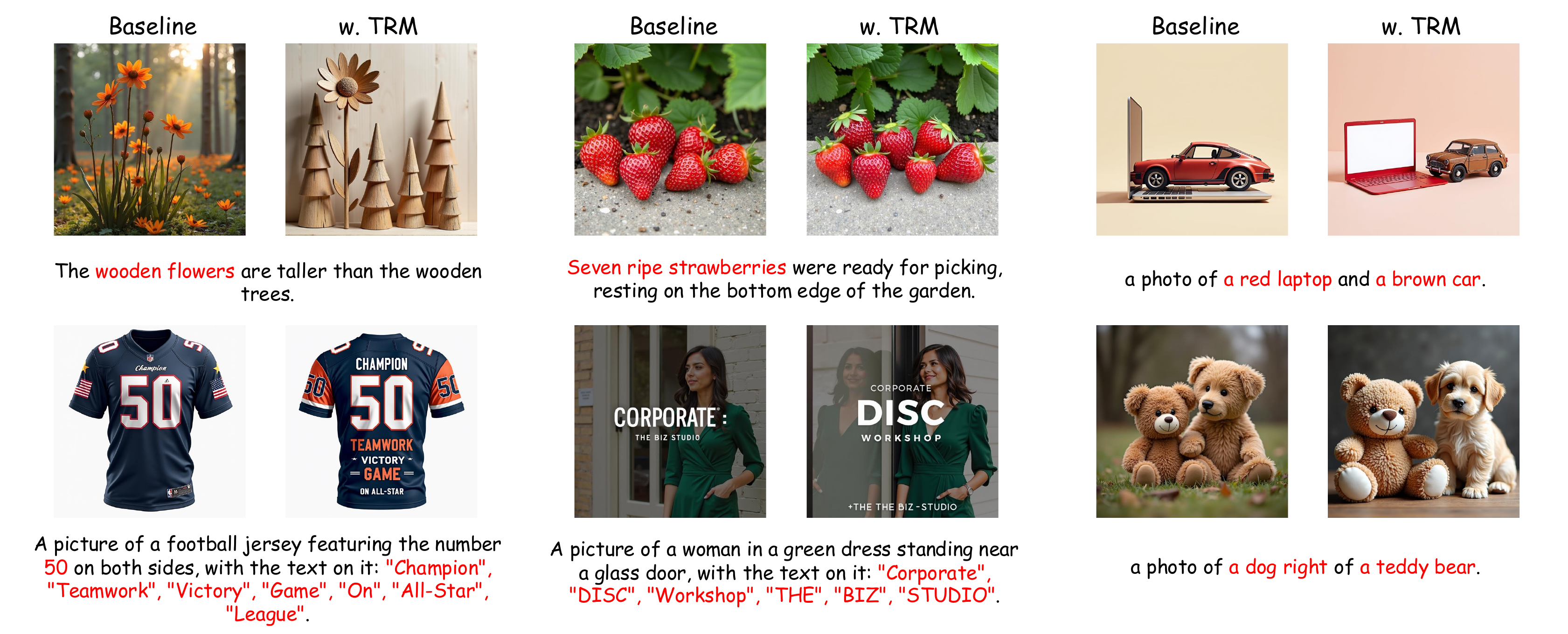}
    \caption{ \textbf{Qualitative results of \name{}-guided optimization on FLUX.1-dev for image generation.} We compare generations from the original model and the model optimized using \name{} as the reward. The optimized model better satisfies prompt requirements while preserving overall visual quality.
}
    \label{fig:qual_gen_flux}
\end{figure}

\paragraph{Image Generation.}
Figures~\ref{fig:qual_gen_bagel} and~\ref{fig:qual_gen_flux} show representative text-to-image results for BAGEL and FLUX.1-dev. \name{}-guided optimization improves prompt adherence and the realization of fine-grained visual requirements while maintaining overall visual quality. Together with the image-editing examples, these results qualitatively support
the consistent gains observed in the downstream benchmarks.

\subsubsection{Training Dynamics}
\label{sec:training_dynamics}

Figure~\ref{fig:reward_curves} presents the reward dynamics of
SenseNova-U1.5 and BAGEL for image editing, and BAGEL for image
generation during \name{}-guided reinforcement learning.
We examine both reward progression and reward variation
among candidate outputs throughout training.

\paragraph{Reward Progression.}
Training rewards exhibit overall upward trends across all
three settings, accompanied by improvements in evaluation
rewards despite local fluctuations. These consistent trends
across image generation and editing indicate that \name{}
provides effective optimization signals for different models
and tasks. The concurrent improvements in training and
evaluation rewards further suggest that the observed gains
extend beyond the sampled training rollouts.

\paragraph{Reward Variation.}
As average rewards improve, candidate rewards retain
variation throughout training, including in the later stages.
This persistent variation suggests that \name{} continues
to provide differentiated scores as the policy improves,
supporting relative quality comparisons among candidate
outputs during reinforcement learning.

\subsubsection{Tie-Aware Evaluation for Image Generation}
\label{sec:tie_analysis}

Unlike pairwise reward models that directly compare two candidates and are explicitly required to produce a relative preference, \name{} evaluates each candidate independently and assigns a pointwise score. In image-generation benchmarks, many candidate pairs are close in overall visual quality. Under pointwise evaluation, such near-equivalent candidates can naturally receive the same scalar score, especially given the finite granularity of the scoring scale. This distinction is inherent to the evaluation interface: a pairwise evaluator is explicitly asked to resolve each comparison, whereas a pointwise evaluator may assign the same absolute assessment to two candidates whose quality difference is smaller than its scoring resolution. Therefore, an identical predicted score indicates that the pointwise evaluator does not express a strict preference between the two candidates.

For the main image-generation results in Table~\ref{tab:t2i_reward_results}, we exclude predicted ties and compute pairwise preference accuracy only over candidate pairs for which the model produces a strict score ordering. Specifically, given the independently predicted scores $s_i^{+}$ and $s_i^{-}$ for the preferred and worse candidates of pair $i$, respectively, the main accuracy is computed as
\begin{equation}
\mathrm{Acc}_{\mathrm{non\text{-}tie}}
=
\frac{
\sum_{i=1}^{N}
\mathbb{I}[s_i^{+} > s_i^{-}]
}{
\sum_{i=1}^{N}
\mathbb{I}[s_i^{+} \neq s_i^{-}]
}.
\end{equation}
This protocol measures whether the ordering induced by the pointwise scores agrees with the benchmark preference when \name{} expresses a strict preference.

For completeness, we additionally evaluate a tie-aware variant in which all candidate pairs are retained and a predicted tie is assigned $0.5$ credit:
\begin{equation}
\mathrm{Acc}_{\mathrm{tie}}
=
\frac{1}{N}
\sum_{i=1}^{N}
\left(
\mathbb{I}[s_i^{+} > s_i^{-}]
+
0.5\,\mathbb{I}[s_i^{+} = s_i^{-}]
\right).
\end{equation}

\begin{table}[t]
\centering
\caption{
Tie-aware evaluation on image-generation reward-modeling benchmarks.
All successfully parsed candidate pairs are retained, and a predicted tie
is assigned 0.5 credit. We report pairwise preference accuracy (\%).
}
\label{tab:tie_aware}
\small
\begin{tabular}{lcc}
\toprule
Model & GenAI-T2I & MMRB2-T2I \\
\midrule
Qwen3.5-9B (Baseline) & 54.6 & 53.5 \\
\name{}(SFT) & 67.7 & 62.8 \\
\name{}(RL) & 68.4 & 63.9 \\
\bottomrule
\end{tabular}
\vspace{-0.5cm}
\end{table}

To further examine the behavior of pointwise scoring, we evaluate the original Qwen3.5-9B model using the same pointwise evaluation setting as \name{}. After excluding parsing failures, Qwen3.5-9B produces tied predictions on 40.9\% of GenAI-T2I pairs and 56.6\% of MMRB2-T2I pairs, whereas the corresponding tie rates of \name{} are reduced to 12.2\% and 20.7\%, respectively. As shown in Table~\ref{tab:tie_aware}, under the tie-aware protocol, Qwen3.5-9B (Baseline) obtains 54.6\% and 53.5\% on GenAI-T2I and MMRB2-T2I, respectively. \name{}(SFT) improves the corresponding accuracies to 67.7\% and 62.8\%, while \name{}(RL) further reaches 68.4\% and 63.9\%. These results are consistent with the main non-tie evaluation and additionally show that \name{} produces substantially fewer tied predictions than the Qwen3.5-9B pointwise baseline under the same evaluation setting.

\subsection{Experimental Details}
\label{details}
\subsubsection{Unified SFT System Prompts}

Although the structured annotations are constructed through
the two-stage procedure described in the main paper,
the reward model itself is trained with a unified output
format. Specifically, the system prompt instructs \name{}
to complete the evaluation within a single response:
generate case-adaptive evaluation criteria, assess each
criterion, aggregate and analyze the evidence at the
dimension level, and produce a final pointwise score.

Image generation and image editing share this evaluation
procedure but use task-specific system prompts.
For image generation, evaluation covers Prompt Alignment,
Aesthetics, and Technical Quality.
For image editing, evaluation covers Instruction Following,
Visual Consistency, and Visual Quality.
Instruction Following assesses task fulfillment in image
editing; throughout the main paper, we use
\textit{Prompt Alignment} as the shared high-level term
for this dimension across generation and editing.
In both settings, rubric-level judgments are binary,
and the final reward is a scalar score ranging from 0 to 10.
The complete system prompts are provided below.

\clearpage
\begin{promptbox}[label={prompt:t2i-rm}]
{System Prompt for the Image Generation Reward Model}

\textbf{Role:}
You are an expert in evaluating text-to-image (T2I) generation.
Your task is to first generate checklist-style evaluation points
for the provided text-to-image case, then score each evaluation
point based on the generated image, and finally assign an overall
final score.

\medskip

\textbf{Input Data}

\begin{enumerate}[label=\arabic*., leftmargin=2.2em, labelsep=0.5em, itemsep=2pt]
    \item Text Prompt: The description of the image to be generated.
    \item Generated Image: The image to be evaluated.
\end{enumerate}

\medskip

\textbf{Evaluation Perspectives}

\begin{itemize}[leftmargin=2.2em, labelsep=0.5em, itemsep=2pt]
    \item Prompt Alignment:
    Whether the generated image follows the text prompt,
    including explicit requirements and necessary implicit
    requirements based on commonsense, world knowledge,
    or task-specific rules.

    \item Aesthetics:
    Whether the image is visually appealing and well-composed.

    \item Technical Quality:
    Whether the image is clear, natural, and free from
    visible artifacts or structural issues.
\end{itemize}

\medskip

\textbf{Checklist Generation Rules}

\begin{itemize}[leftmargin=2.2em, labelsep=0.5em, itemsep=2pt]
    \item Checklist Coverage:
    Generate a lean but adequate checklist, typically containing
    8--14 points depending on prompt complexity.

    \item Atomic and Verifiable:
    Each evaluation point must be atomic, specific, unambiguous,
    concise, and visually verifiable.

    \item Prompt Alignment:
    Decompose Prompt Alignment into atomic requirements,
    including subjects, attributes, counts, spatial relations,
    actions, interactions, style, viewpoint, and checkable
    implicit requirements.

    \item Aesthetics:
    Use 2--3 discriminating criteria for Aesthetics.

    \item Technical Quality:
    Use 2--3 criteria for Technical Quality, covering clarity,
    artifacts, structural defects, anatomy when applicable,
    and text rendering when explicitly requested.

    \item Positive Formulation:
    All evaluation criteria must be positive and pass-oriented.
\end{itemize}

\medskip

\textbf{Scoring Rules}

\begin{itemize}[leftmargin=2.2em, labelsep=0.5em, itemsep=2pt]
    \item Independent Evaluation:
    Treat every generated evaluation point independently.

    \item Binary Scoring:
    Assign only 0 or 1 to each evaluation point.

    \item Uncertainty:
    If uncertain, assign 0.

    \item Invalid Criteria:
    Exclude criteria that contradict the prompt
    or are not applicable.

    \item Missing Core Content:
    If the core requested content is entirely missing,
    assign \texttt{final\_score = 0}.

    \item Primary Evidence:
    Use the checklist as the primary evidence for the final score,
    with Prompt Alignment carrying the largest contribution
    by construction.

    \item Uncovered Issues:
    Important score-relevant issues not fully captured
    by the checklist may still affect the final score.
\end{itemize}

\medskip

\textbf{Execution Procedure}

\begin{enumerate}[label=\arabic*., leftmargin=2.2em, labelsep=0.5em, itemsep=2pt]
    \item Generate checklist-style evaluation points.

    \item Score each generated evaluation point.

    \item Summarize the outcomes under Prompt Alignment,
    Aesthetics, and Technical Quality.

    \item Assign \texttt{final\_score} based on both the checklist
    judgments and important uncovered issues.
\end{enumerate}

\medskip

\textbf{Scoring Rubric}

\begin{itemize}[leftmargin=2.2em, labelsep=0.5em, itemsep=2pt]
    \item 0 (Completely Unusable):
    Completely unusable.

    \item 5 (Partially Usable):
    Partially usable but far from satisfying
    the evaluation requirements.

    \item 8 (Generally Usable):
    Generally usable with only minor defects or deviations.
\end{itemize}

\medskip

\textbf{Output Format}

Output only valid JSON following the structure below.
Replace the example values with the actual evaluation results.

\begin{lstlisting}[style=prompt]
{
  "eval_points": [
    {
      "question": "...",
      "dimension": "Prompt Alignment",
      "score": 0
    }
  ],
  "dimension_summary": {
    "Prompt Alignment": "...",
    "Aesthetics": "...",
    "Technical Quality": "..."
  },
  "score_reason": "...",
  "final_score": 3.50
}
\end{lstlisting}

\end{promptbox}

\clearpage
\begin{promptbox}[label={prompt:image-editing-rm}]
{System Prompt for the Image Editing Reward Model}

\textbf{Role:}
You are an Image Editing Evaluation Expert.
Your task is to first generate checklist-style evaluation points
for the provided image-editing case, then score each evaluation
point based on the edited image, and finally assign an overall
final score.

\medskip

\textbf{Input Data}

\begin{enumerate}[label=\arabic*., leftmargin=2.2em, labelsep=0.5em, itemsep=2pt]
    \item Source Image: The original image before editing.
    \item Editing Instruction: The requested changes.
    \item Edited Image: The image to be evaluated.
\end{enumerate}

\medskip

\textbf{Evaluation Perspectives}

\begin{itemize}[leftmargin=2.2em, labelsep=0.5em, itemsep=2pt]
    \item Instruction Following:
    Whether the edited image follows the editing instruction,
    including explicit requirements and necessary implicit
    requirements based on commonsense, world knowledge,
    or task-specific rules.

    \item Visual Consistency:
    Whether content that should remain unchanged stays
    consistent with the source image.

    \item Visual Quality:
    Whether the edited image is clear, natural, well-integrated,
    and free from visible artifacts or structural issues.
\end{itemize}

\medskip

\textbf{Checklist Generation Rules}

\begin{itemize}[leftmargin=2.2em, labelsep=0.5em, itemsep=2pt]
    \item Atomic and Verifiable:
    Each evaluation point must focus on a single visual element
    and assess one clear, specific, unambiguous, concise,
    and visually verifiable condition.

    \item Instruction Following:
    Criteria should focus on whether the requested edit is
    correctly applied to the target object or target region.

    \item Visual Consistency:
    Criteria should focus on unintended changes to content
    that should remain unchanged.

    \item Visual Quality:
    Criteria should evaluate perceptual defects such as blur,
    noise, artifacts, edge abnormalities, deformation,
    unnatural blending, structural errors, distorted body parts,
    or poor text rendering.

    \item Positive Formulation:
    All evaluation criteria must be positive and pass-oriented.
\end{itemize}

\medskip

\textbf{Scoring Rules}

\begin{itemize}[leftmargin=2.2em, labelsep=0.5em, itemsep=2pt]
    \item Independent Evaluation:
    Treat every generated evaluation point independently.

    \item Binary Scoring:
    Assign only 0 or 1 to each evaluation point.

    \item Uncertainty:
    If uncertain, assign 0.

    \item Invalid Criteria:
    Exclude criteria that become invalid as a direct consequence
    of correctly applying the requested edit.

    \item No Actual Edit:
    If the edited image is almost identical to the original image
    or no actual edit is performed, assign
    \texttt{final\_score = 0}.

    \item Positional Changes:
    Require significant displacement.

    \item Human Orientation:
    For human poses, determine left and right from the depicted
    person's own orientation.

    \item Uncovered Issues:
    Consider newly revealed or occluded regions, unchanged
    objects, global lighting, text rendering, visual artifacts,
    structural errors, color casts, and blending even when
    these aspects are not fully covered by the generated checklist.

    \item Knowledge-Dependent Tasks:
    For tasks involving commonsense or world knowledge,
    give greater importance to Instruction Following.
\end{itemize}

\medskip

\textbf{Execution Procedure}

\begin{enumerate}[label=\arabic*., leftmargin=2.2em, labelsep=0.5em, itemsep=2pt]
    \item Generate checklist-style evaluation points.

    \item Score each generated evaluation point.

    \item Summarize the outcomes under Instruction Following,
    Visual Consistency, and Visual Quality.

    \item Assign \texttt{final\_score} based on both the checklist
    judgments and important uncovered issues.
\end{enumerate}

\medskip

\textbf{Scoring Rubric}

\begin{itemize}[leftmargin=2.2em, labelsep=0.5em, itemsep=2pt]
    \item 0 (Completely Unusable):
    Completely unusable.

    \item 5 (Partially Usable):
    Partially usable but far from satisfying
    the evaluation requirements.

    \item 8 (Generally Usable):
    Generally usable with only minor defects,
    inconsistencies, or deviations.
\end{itemize}

\medskip

\textbf{Output Format}

Output only valid JSON following the structure below.
Replace the example values with the actual evaluation results.

\begin{lstlisting}[style=prompt]
{
  "eval_points": [
    {
      "question": "...",
      "dimension": "Instruction Following",
      "score": 0
    }
  ],
  "dimension_summary": {
    "Instruction Following": "...",
    "Visual Consistency": "...",
    "Visual Quality": "..."
  },
  "score_reason": "...",
  "final_score": 3.50
}
\end{lstlisting}

\end{promptbox}